# Victoria Amazonica Optimization (VAO)

# An Algorithm Inspired by the Giant Water Lily Plant

**Seyed Muhammad Hossein Mousavi\***
*Developer at Pars AI Company*
*Tehran, Iran*
*ORCID: 0000-0001-6906-2152*
*mosavi.a.i.buali@gmail.com*

*Abstract*

The Victoria Amazonica plant, often known as the Giant Water Lily, has the largest floating spherical leaf in the world, with a maximum leaf diameter of 3 meters. It spreads its leaves by the force of its spines and creates a large shadow underneath, killing any plants that require sunlight. These water tyrants use their formidable spines to compel each other to the surface and increase their strength to grab more space from the surface. As they spread throughout the pond or basin, with the earliest-growing leaves having more room to grow, each leaf gains a unique size. Its flowers are transsexual and when they bloom, Cyclocephala beetles are responsible for the pollination process, being attracted to the scent of the female flower. After entering the flower, the beetle becomes covered with pollen and transfers it to another flower for fertilization. After the beetle leaves, the flower turns into a male and changes color from white to pink. The male flower dies and sinks into the water, releasing its seed to help create a new generation. In this paper, the mathematical life cycle of this magnificent plant is introduced, and each leaf and blossom are treated as a single entity. The proposed bio-inspired algorithm is tested with 24 benchmark optimization test functions, such as Ackley, and compared to ten other famous algorithms, including the Genetic Algorithm. The proposed algorithm is tested on 10 optimization problems: Minimum Spanning Tree, Hub Location Allocation, Quadratic Assignment, Clustering, Feature Selection, Regression, Economic Dispatching, Parallel Machine Scheduling, Color Quantization, and Image Segmentation and compared to traditional and bio-inspired algorithms. Overall, the performance of the algorithm in all tasks is satisfactory.

## 1. INTRODUCTION

Nature provides everything we require, including solutions to the most intricate problems. Metaheuristics Optimization (Bianchi et al., 2009) refers to problem-solving techniques based on biology in nature or bio-inspired behavior of animals, plants, or natural phenomena, which aim to find the most optimal solution to a complex mathematical problem. Occasionally, these algorithms are referred to as nature-inspired algorithms (Mousavi et al., 2017a; Vikhar, 2016), but for the sake of clarity, the term 'bio-inspired' algorithms (Darwish, 2018) will be used throughout the paper.

*Corresponding Author: Seyed Muhammad Hossein Mousavi (+98-09332892726)*

Due to their iterations and generations, running these algorithms on older devices was extremely time-consuming due to their high runtime complexity; however, modern personal computers can handle them easily. They can be used as single-objective or multi-objective optimization problems in a variety of areas, from agriculture to medicine (Del Ser et al., 2019) and even in subfields of computer science (particularly Artificial Intelligence (AI) (Flasiński, 2016)) such as machine learning (Mousavi et al., 2020), and so on. One of the earliest implementations of these algorithms dates back to 1960 by Bienert, Rechenberg, and Schwefel, who created Evolution Strategies (ES) and developed the 2002 algorithm (Beyer & Schwefel, 2002). Since then, numerous nature-inspired algorithms for optimization tasks have been developed, the most well-known of which are the three algorithms of Genetic Algorithm (GA) (Mitchell, 1998), Particle Swarm Optimization (PSO) (Kennedy & Eberhart, 1995), and Ant Colony Optimization (ACO) (Dorigo et al., 2006). Each of these mentioned algorithms has its advantages and disadvantages which this research tries to introduce an algorithm with a higher level of advantages compared to other similar ones. Their advantages and disadvantages return to their performance quality on assigned optimization tasks or problems to solve such as Regression, Clustering, Feature Selection, Hub Location Allocations, Segmentation, Minimum Spanning Tree and so on which proposed algorithms show reliable performance in all of them compare to others.

Victoria Amazonica, also known as Amazon Water Lily, Amazon Water-platter, Giant Water Lily, and Royal Water Lily, is the largest species of Nymphaeaceae water lily (Prance & Arias, 1975). It was discovered in Bolivia in 1801 by the Czech botanist Tadeas Haenke (Holway, 2013). In 1837, however, the English botanist John Lindley created the genus Victoria and named the species Victoria Regia after Queen Victoria (Holway, 2013). It is native to South America (Brazil, Guyana, Peru, Colombia, and Venezuela) and the shallow waters of the Amazon River basin, with a stem length of 8 meters, a flower diameter of 40 centimeters, and a leaf size of 3 meters in diameter, that can hold up to 60 kilograms of weight (Holway, 2013; Prance & Arias, 1975). There are veins beneath the leaf, which create a space between the water and the leaf's main body, which is the primary factor in its ability to support such weight. The spiny seed grows from the bottom of the pond, basin, or river. After reaching the surface, it begins to use its bud to clear the surface. Following this, the leaf begins to dominate the surface by growing and spreading 20 centimeters per day. By enlarging the leaf, those plants beneath it would gradually perish due to insufficient access to sunlight. However, surface plants, including small water lilies, will be crushed and punctured by the growing force and spun by the fittest (BBC, 2022; Glimn-Lacy & Kaufman, 2006; Holway, 2013). They are naturally competitive for space and do not enjoy sharing it. Leaves and flowers are viewed as generational individuals.

Scarab beetles of the genus Cyclocephala (Maia & Schlindwein, 2006; Seymour & Matthews, 2006) are responsible for the flower's pollination process. On the first night of pollination, the white female flower begins to slowly open and emit a pineapple-like scent. Inside, the temperature is warmer and there are abundant nectars for beetles. After a beetle enters a flower, the flower closes at dawn, trapping the insect inside. The beetle transports pollen from another male flower to the white female flower to fertilize it. The second night, the female flower transforms into a male, turns purple, and covers the beetle with its own pollen. Pollen is transferred by a beetle to a white female flower for fertilization. Finally, the male pink flower dies and sinks into the water, allowing the precious seed to germinate and start a new generation (Holway, 2013; Les et al., 1999). This plant belongs to the Nymphaeaceae or water lily family (Les et al., 1999). Most water lilies are capable of hybridization. The combination of two species of the genus Victoria, Victoria Amazonica and Victoria Cruziana, is one example (Lamprecht et al., 2002). This variant is known as "The Victoria Longwood Hybrid" (Jian et al., 2010; NCSU, n/a; Taylor et al., 2013; Zhang et al., 2021). It exhibits the best qualities of both parents. These are in large size and deep red color of the leaf undersides of Victoria Amazonica, as well as the large lip and relative hardiness of Victoria Cruziana. This variant is regarded as a success and a beneficial mutation. However, this mutation could result in a weaker plant, which is the mutation's disadvantage. We considered a pond as the setting, and a variety of water lily species were involved. This indicates that various species of water lilies could be pollinated by insects. This plant is a decent source of inspiration as its life cycle has a diverse success rate in each generation and together, they illustrate a swarm-like behavior which combined could be used as an optimization technique. Figure 1 depicts the life cycle of the giant water lily plant, as described previously, and Figure 2 depicts a pond of giant water lilies and the diameter ($\varnothing$) of each leaf's spreading power.

This paper is organized into four sections. Section one is devoted to fundamentals and basics. Some of the relevant research conducted by other researchers in the field of bio-inspired algorithms is covered in the second section. The third section provides a detailed description of the proposed method, followed by validations, results, and comparisons with other methods. The fourth section contains a conclusion and suggestions for future work. The following box shows terms and abbreviations which are used throughout the paper.

| | |
|---|---|
| Victoria Amazonica Optimization Algorithm (VAO) | Quadratic Assignment Problem (QAP) |
| Genetic Algorithm (GA) | Parallel Machine Scheduling (PMS) |
| Firefly Algorithm (FA) | Economic Dispatching (ED) |
| Particle Swarm Optimization Algorithm (PSO) | Image Quantization (IQ) |
| Ant Colony Optimization (ACO) | Image Segmentation (IS) |
| Artificial Bee Colonies Algorithm (ABC) | Gaussian Mixture Model (GMM) |
| Evolution Strategies Algorithm (ES) | K-Nearest Neighbor (K-NN) |
| Imperialist Competitive Algorithm (ICA) | Support Vector Machine (SVM) |
| Biogeography-Based Optimization Algorithm (BBO) | Feature Selection (FS) |
| Teaching Learning Based Optimization Algorithm (TLBO) | Number of Features (NF) |
| Galaxy Gravity Optimization Algorithm (GGO) | Means Square Error (MSE) |
| Grey Wolf Optimizer Algorithm (GWO) | Root Mean Square Error (RMSE) |
| African Vultures Optimization Algorithm (AVOA) | Correlation Coefficient (CC) |
| Harmony Search (HS) | Principal Component Analysis (PCA) |
| Differential Evolution (DE) | Fuzzy C-Means (FCM) |
| Bee-eater Hunting Strategy (BEH) | 2-Dimension (2-D) |
| Weevil Damage Optimization Algorithm (WDOA) | 3-Dimension (3-D) |
| Artificial Intelligence (AI) | Non-deterministic Polynomial-time Hardness (NP-Hardness) |
| Standard Deviation (std) | One Step Secant (OSS) |
| Expansion ($\varnothing$) | Power Total (PT) |
| Intra Competition ($\lambda$) | Power Loss (PL) |
| Mutation ($\mu$) | Power Demand (PD) |
| Drawback coefficient 1 ($\omega$) | Matthews Correlation Coefficient (MCC) |
| Drawback coefficient 2 ($\psi$) | Peak Signal to Noise Ratio (PSNR) |
| Hub Location Allocation (HLA) | Structural Similarity Index Measure (SSIM) |
| Minimum Spanning Tree (MST) | Ground Truth (GT) |

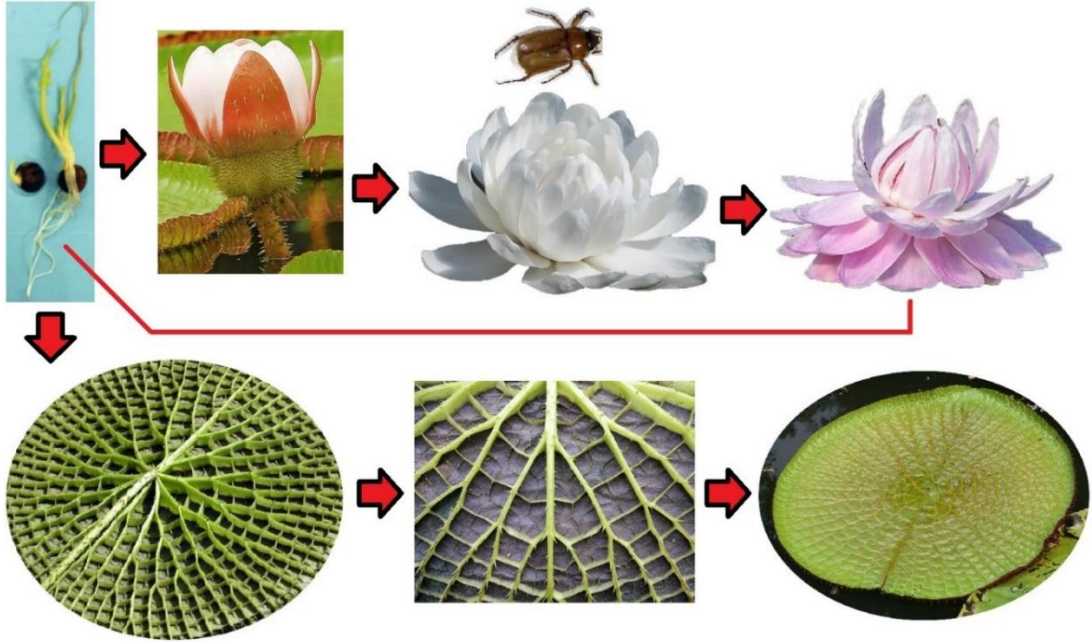

*Figure 1.* Life cycle of the giant water lily (Victoria Amazonica).

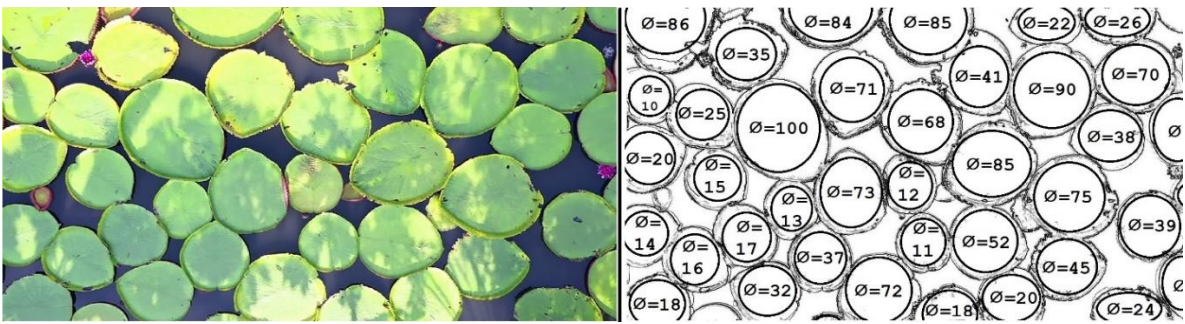

*Figure 2.* Left: Pond of giant water lilies; Right: distribution or spreading the power of each leaf in diameter (⌀), with values between 10 and 100, as an example of -dimensional optimization problem.

## 2. PRIOR RELATED WORKS

The Genetic Algorithm (GA) (Mitchell, 1998) is one of the most well-known evolutionary algorithms. Using Charles Darwin's theory of natural selection searches heuristically for the optimal solutions and individuals. Crossover, selection, and mutation serve as its foundation. Particle Swarm Optimization (PSO) is another well-known bio-inspired optimization technique (Kennedy & Eberhart, 1995). Each individual has a position and velocity in order to interact with other particles in the swarm and find a better solution. The social behavior of birds and fishes toward the best position creates the base for it. Designing objective functions for GA is hard and PSO easily falls in local minima in high dimensional spaces. Other than that, both of them are computationally expensive and time-consuming algorithms.

The Differential Evolution (DE) algorithm (Storn et al., 1997) is one of the most popular and effective metaheuristic algorithms which falls into the evolutionary algorithms category. DE was created in the mid-1990s and was designed to optimize problems over continuous domains. This approach originated from Kenneth's Price attempts to solve the Tchebycheff Polynomial fitting Problem that had been posed to him by Rainer Storn.

Ant Colony Optimization (ACO) (Dorigo et al., 2006) is yet another reputable swarm-based nature-inspired algorithm for single and multi-objective optimization problems. Through iterations and generations, artificial ants collaborate to find the optimal path to a destination. The exploration, exploitation rate, and convergence speed of ACO is its main disadvantage. The Artificial Bee Colony (ABC) (Karaboga, 2005) algorithm implements the social behavior of a honey bee swarm. This algorithm focuses on the interaction of worker bees, onlookers, and scouts with the hive and food sources in order to find the optimal solution. This algorithm suffers from premature convergence in new periods.

The Imperialist Competitive Algorithm (ICA) (Atashpaz-Gargari & Lucas, 2007) refers to one of the algorithms based on the political strategy of empires to acquire more colonies and falls into the human-based category. Empires attempt to expand their colonies through Competition, Assimilation, and Revolution operators, of which mutation is the Revolution operator. Those who gain the most colonies are candidates for the best solutions. However, this algorithm suffers from high complexity.

The Biogeography-Based Optimization (BBO) algorithm (Simon, 2008) is a swarm-based nature-inspired robust algorithm for various optimization problems. This algorithm is all about transferring living organisms from one habitat to another which provides them with better living conditions and more room to grow. Immigration, emigration, and the human suitability index are used to determine the best place for newcomers to live.

The Firefly algorithm (FA) (Yang, 2010) is yet another efficient swarm-based optimization algorithm. FA consists of five primary components: population size (fireflies), the light intensity of each firefly, coefficient of light absorption, coefficient of attraction, and mutation rate. It works by relocating fireflies with a lower light intensity to those with a higher intensity. There is a high probability of being trapped in local optima because FA is a local search algorithm. Teaching Learning Based Optimization (TLBO) (Rao, 2016) is one of the most recent and powerful optimization algorithms. The TLBO algorithm consists of two primary phases, teacher and learner, with the teacher phase occurring first. It begins with determining the mean of each designed variable and determining the optimal solution for a teacher. After evaluation, the teacher has deemed the best solution for the entire population. In the learner phase, the teacher selected in the previous step shares his or her knowledge with the students in order to raise the class's average level of knowledge.

In 2017, an algorithm inspired by outer space emerged called the Galaxy Gravity Optimization (GGO) algorithm (Mousavi et al., 2017a) and falls into the physique-based category. The algorithm is based on the life cycle of comets, specifically the Kuiper belt. It is all about how their size increases due to gravitational force and velocity, and how they lose mass or weight during their orbital journey. After multiple perigees, the candidates with the greatest mass are deemed to be the best options.

Another great example of these swarm-based nature-inspired algorithms is called Gray Wolf Optimizer Algorithm (GWO) (Mirjalili et al., 2014). This algorithm simulates the leadership hierarchy and hunting strategy of wolf packs which consisted of four grey wolves of alpha, beta, omega, and delta are the main assets of this algorithm for the whole process. Drawbacks are low solving accuracy, bad local searching ability, and slow convergence rate.

One of the most recent swarm-based algorithms in metaheuristics is called the African Vulture Optimization Algorithm (AVOA) (Abdollahzadeh et al., 2021) which is inspired by the exploring and navigation power of African vultures. Its main task is to solve global optimization problems. The only problem here is the high computational time.

A mentionable physique-based optimization algorithm is Harmony Search (HS) (Geem et al., 2001) which mimics musicians producing music. The algorithm starts with generating some random solutions in the harmony memory and evaluating them. New populations or harmonies generates by harmony memory consideration rate factor and pitch adjustment rate which is mutation takes place. The goal is to find the best harmonies among the population which are possible solution candidates. It is a fast algorithm but takes more time and iterations to converge.

Another novel swarm-based nature-inspired algorithm is called Bee-eater Hunting Strategy (BEH) algorithm (Mousavi, 2022) which is inspired by the bee-eater hunting bee. There are two operators of peak power and adjustment power in the BEH algorithm which aids bee-eater to navigate and attack bees in the air. The mutation is the fight over a big prey and speed is its advantage. Weevil Damage Optimization Algorithm (WDOA) is introduced by Mousavi and Mirinezhad in 2022 (Mousavi and Mirinezhad, 2022). It is a swarm-based algorithm for optimization purposes and is inspired by Weevil's damage on crops and cereal grains which this damage is called Damage Decision Variable or DDV. The cost function is based on the amount of damage Weevil can make and that relies on the good environment for reproduction which this environment is considered as Environmental Situation Index or ESI. Each weevil individual has snout and fly power respectively and mutation is called Reproduction Environment Rate or RER. WDOA is a fast and robust algorithm but it will trap in local optima.

Multiple types of nature-inspired algorithms, including DNA biology, fish and bird swarming, the social behavior of flying and on-the-ground insects, political relations between governments, biogeography behavior of animals, and even school social behavior, were discussed in this section. The algorithm proposed in this paper is based on the social behavior of giant water lily plants and is a novel attempt in the literature.

## 3. VICTORIA AMAZONICA OPTIMISATION (VAO) ALGORITHM

As stated in the introduction, the VAO algorithm is concerned with the distribution of the initial population (Leaves and Flowers) and their corresponding spreading power or "Expansion" on the surface. This algorithm is mainly a swarm local search-based metaheuristic algorithm, so the only drawback is a chance of trapping in local optima, other than that it is fast and robust for a lot of optimization tasks. In this study, the scientific term for diameter "∅" illustrates their circular growth as they expand in circular form. This expansion is followed by the amount of space they can obtain by forcibly displacing one another with their rising strength and spines. This competition is referred to as "Intra Competition" or "λ" for formulating purposes.

Additionally, there are three frequent difficulties that hinder plant growth. These are the death of beetles within the flower, improper or no pollination by beetles, and a decrease in temperature. All of these elements could be detrimental to the process, and we refer to them all as "ω" here. Clearly, a greater value of ω indicates a weaker plant. Pests, such as water lily Aphids, can also harm the plant by eating its leaves and creating holes in them. The "ψ" symbol represents this threat value in the paper. The lower the value of "ψ", the more favorable the conditions for plant growth and expansion.

Finally, the mutation occurs when beetles in the pond pollinate the flower with another type of water lily. This phenomenon is referred to as "Hybrid Mutation" and is represented by the "μ" symbol. As stated in the introduction, this mutation can occur in both positive and negative directions, at a rate of 0.2% for every generation. The largest and strongest leaf is the best or "α". Additionally, the VAO pseudo code is as below, while Figure 3 shows the flowchart of the VAO method.

$$VAO = \sum_{i=1}^{n}\sum_{j=1}^{n}(xij\ [\varnothing ij, \lambda ij] + \psi + \omega) * (\mu) \tag{1}$$

| **VAO Pseudo Code** |
| --- |
| *Start* |
| *Generating population of plants $x_i$ (i=1,2,…,n)* |
| *Define Expansion $\varnothing_i$ in $x_i$* |
| *Define Intra Competition $\lambda_i$ in $x_i$* |
| *Define Drawback coefficient of ω in $x_i$(random range in [0.1 to 0.3])* |
| *Define Drawback coefficient of ψ in $x_i$(random range in [0.1 to 0.3])* |
| *Define Hybrid Mutation Rate of μ = 0.2* |
|    *While maximum iterations are not satisfied* |
|       *For i=1 to n plants* |
|          *For j=1 to n plants (Intra Competition)* |
|          *If ($\varnothing_i > \varnothing_j$ or $\lambda_i > \lambda_j$) for xi (i=1,2,…,n)* |
|             *plant i goes plant j* |
|          *End if* |
|            *Apply hybrid mutation μ* |
|            *Apply Drawback coefficient ω and ψ* |
|            *Evaluate new solutions by cost function and update Expansion* |
|         *End* |
|       *End* |
|       *Sort and rank plants and find the current global best (α)* |
|       *Generating new generation* |
|    *End of while* |
| *End* |

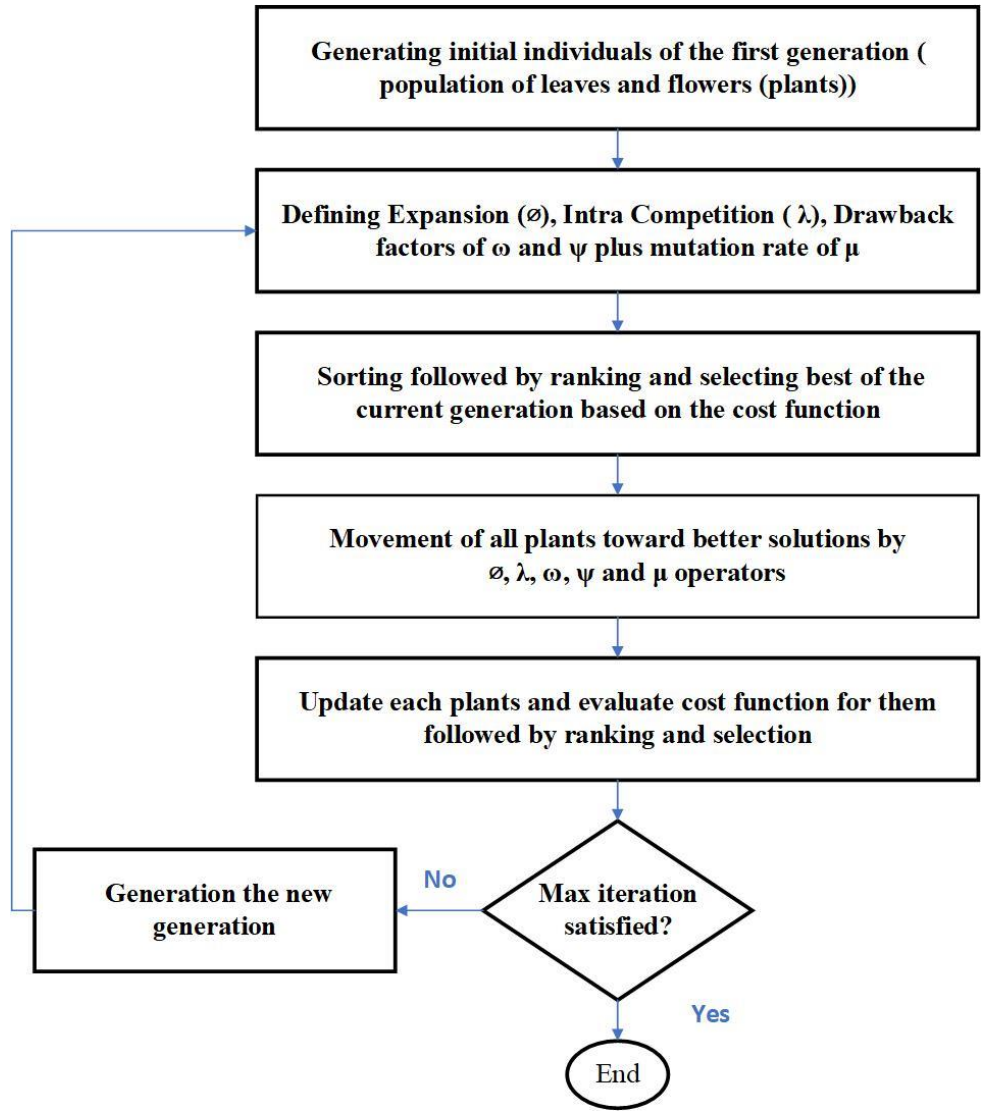

*Figure 3.* Flowchart of the proposed Victoria Amazonica Optimization (VAO) algorithm.

## 4. EVALUATION AND RESULTS

Various academics have proposed performance test functions or artificial landscapes (Back, 1996) over the years to evaluate the performance of bio-inspired algorithms. These algorithms evaluate the evolutionary algorithm's robustness, precision, convergence rate, and overall performance. They consist mostly of single-objective, constraint-type, and multi-objective optimization forms (Back, 1996; Momin *et al.*, 2013), as well as a variety of landscape shapes designed to trap the algorithm in iterations. If the algorithm can pass all of these performance functions without becoming trapped in local minima\maxima (Weir, 2008), then it is deemed a suitable algorithm for optimization. 24 performance test functions of Ackley (Back, 1996; Momin *et al.*, 2013), Powell (Laguna & Marti, 2005; Momin *et al.*, 2013), Rastrigin (Pohlheim, 2006; Vanaret, 2020), Pyramid (Mousavi, 2018), Booth (Momin *et al.*, 2013), Zakharov (Momin *et al.*, 2013), De Jong (Momin *et al.*, 2013), Michalewicz (Molga & Smutnicki, 2005), and Beale (Momin *et al.*, 2013),

Dixon, Bukin 6, Cross-in-tray, Griewank, Bohochevsky, Gold stein, Matyas, Branin, Schwefel, Rosenbrock, Trid, Egg holder, Easom (Momin *et al*., 2013), Levy (Jamil et al., 2013), and Bird (Mishra., 2006) are utilized to evaluate the proposed VAO algorithm. Table 1 presents the characteristics of the functions. In addition, Figure 4 depicts the three-dimensional form of the optimization functions utilized in the experiment. One experiment using the VAO algorithm with 20 variables in the range of [-10, 10], during 120 iterations with 20 plants as populations, is conducted on the functions in Table 1 and presented in Figure 5.

Table 1

*Optimization Test Functions*

| NAME | EQUATION | GLOBAL MINIMUM | RANGE |
|---|---|---|---|
| ACKLEY – MULTIMODAL – MANY LOCAL MINIMA | $f1(\mathbf{x}) = -20e^{\left(-0.02\sqrt{D^{-1}\sum_{i=1}^{D}x_i^2}\right)} - e^{\left(D^{-1}\sum_{i=1}^{D}\cos{(2\pi x_i)}\right)} + 20 + e$ | $\mathbf{x}^* = (0,...,0), f(\mathbf{x}^*) = 0$ | $-35 \leq x_i \leq 35$ |
| POWELL – UNIMODAL – VALLEY-SHAPED | $f2(\mathbf{x}) = \sum_{i=1}^{D/4}(x_{4i-3} + 10x_{4i-2})^2 + 5(x_{4i-1} + x_{4i})^2 + (x_{4i-2} + x_{4i-1})^4 + 10(x_{4i-3} + x_{4i})^4$ | $\mathbf{x}^* = (3,-1,0,1,...,3,-1,0,1), f(\mathbf{x}^*) = 0$ | $-4 \leq x_i \leq 5$ |
| RASTRIGIN – MULTIMODAL – MANY LOCAL MINIMA | $f3(\mathbf{x}) = -20\exp\left(-0.5\sqrt{-0.2(x^2+y^2)}\right) - \exp(0.5(\cos(2\pi y)))$ | $\mathbf{x}^* = (0,...,0), f(\mathbf{x}^*) = 0$ | $-5 \leq x_i \leq 5$ |
| PYRAMID-UNIMODAL-VALLEY-SHAPED | $f4(\mathbf{x}) = \partial(sin(p .* 2)) * 12 + cot(\sqrt{(exp^\pi)}) * \pi. * (\frac{\sqrt{0.1}}{\pi} * ((\frac{\sqrt{sqrt}(\pi+12)}{20})) - 20 * exp^{\left(-\frac{1}{73}*exp\sqrt{\frac{12}{(length(p.*313))}}\right)}$ | $\mathbf{x}^* = (0,...,0), f(\mathbf{x}^*) = 0$ | $-40 \leq x_i \leq 40$ |
| BOOTH MULTIMODAL – PLATE-SHAPED | $f5(\mathbf{x}) = (x_1 + 2x_2 - 7)^2 + (2x_1 + x_2 - 5)^2$ | $\mathbf{x}^* = (1,3), f(\mathbf{x}^*) = 0$ | $-10 \leq x_i \leq 10$ |
| ZAKHAROV - UNIMODAL – PLATE-SHAPED | $f6(\mathbf{x}) = \sum_{i=1}^{D}x_i^2 + \left(\frac{1}{2}\sum_{i=1}^{D}ix_i\right)^2 + \left(\frac{1}{2}\sum_{i=1}^{D}ix_i\right)^4$ | $\mathbf{x}^* = (0,...,0), f(\mathbf{x}^*) = 0$ | $-5 \leq x_i \leq 5$ |
| DE JONG – UNIMODAL – PLATE-SHAPED | $f7(\mathbf{x}) = \sum_{i=1}^{D}x_i^2$ | $\mathbf{x}^* = (0,...,0), f(\mathbf{x}^*) = 0$ | $-10 \leq x_i \leq 10$ |
| MICHALEWICZ – MULTIMODAL – STEEP RIDGES | $f8(\mathbf{x}) = -\sum_{i=1}^{D}\sin{(x_i)}\left(\sin{\left(\frac{ix_i^2}{\pi}\right)}\right)^{2m}$ | $\mathbf{x}^* = (2.203191.57049),1.57049), f(=-1.8013$ for $n = 2$ | $0 \leq x_i \leq \pi, m = 10$ |
| BEALE – UNIMODAL – PLATE-SHAPED | $f9(\mathbf{x}) = (1.5 - x_1 - x_1x_2)^2 + (2.25 - x_1 - x_1x_2^2)^2 + (2.625 - x_1 - x_1x_2^3)^2$ | $\mathbf{x}^* = (3,0.5), f(\mathbf{x}^*) = 0$ | $-4.5 \leq x_i \leq 4.5$ |
| MATYAS UNIMODAL – PLATE-SHAPED | $f10(\mathbf{x}) = 0.26(x_1^2 + x_2^2) - 0.48x_1x_2$ | $\mathbf{x}^* = (0,...,0), f(\mathbf{x}^*) = 0$ | $-10 \leq x_i \leq 10$ |
| TRID UNIMODAL – BOWL-SHAPED | $f11(\mathbf{x}) = \sum_{i=1}^{D}(x_i - 1)^2 - \sum_{i=2}^{D}x_ix_{i-1}$ | $f(\mathbf{x}^*) = -50$ | $-36 \leq x_i \leq 36$ |
| SCHWEFEL - EASOM UNIMODAL – STEEP RIDGES | $f12(\mathbf{x}) = -\sum_{i=1}^{D}|x_i| + \prod_{i=1}^{D}|x_i|$ $f13(\mathbf{x}) = -\cos{(x_1)}\cos{(x_2)}e^{[-(x_1-\pi)^2-(x_2-\pi)^2]}$ | $\mathbf{x}^* = (0,...,0), f(\mathbf{x}^*) = 0$ $\mathbf{x}^* = (\pi,\pi), f(\mathbf{x}^*) = 0$ | $-10 \leq x_i \leq 10$ $-100 \leq x_i \leq 100$ |

| | | | |
|---|---|---|---|
| **ROSENBROCK – UNIMODAL – VALLY SHAPED** | $f14(\mathbf{x}) = \sum_{i=1}^{D} [100(x_{i+1} - x_i^2)^2 + (x_i - 1)^2]$ | $\mathbf{x}^* = (1, \dots, 1), f(\mathbf{x}^*) = 0$ | $-30 \leq x_i \leq 30$ |
| **BOHACHEVSKY MULTIMODAL – BOWL SHAPED** | $f15(\mathbf{x}) = x_1^2 + 2x_2^2 - 0.3\cos(3\pi x_1) - 0.4\cos(4\pi x_2) + 0.7$ | $\mathbf{x}^* = (0, \dots, 0), f(\mathbf{x}^*) = 0$ | $-100 \leq x_i \leq 100$ |
| **BUKIN 6 – MULTIMODAL -** | $f16(\mathbf{x}) = 100\sqrt{\|x_2 - 0.01x_1^2\|} + 0.01\|x_1 + 10\|$ | $\mathbf{x}^* = (-10,0), f(\mathbf{x}^*) = 0$ | $-15 \leq x_1 \leq -5$ $-3 \leq x_2 \leq 3$ |
| **BRANIN – MULTIMODAL – UNEQUAL SHAPED** | $f17(\mathbf{x}) = \left(x_2 - \frac{5.1}{4\pi^2}x_1^2 + \frac{5.1}{\pi}x_1 - 6\right)^2 + 10\left(1 - \frac{1}{8\pi}\right)\cos(x_1) + 10$ | $\mathbf{x}^* = (\{-3.142, 12.275\}, \{3.142, 2.275\} = 0.398$ | $-5 \leq x_1 \leq 10$ $0 \leq x_2 \leq 15$ |
| **EGG HOLDER – MULTIMODAL – MANY LOCAL MINIMA** | $f18(\mathbf{x}) = \sum_{i=1}^{D-1} \left[ -(x_i + 47)\sin\sqrt{\left|x_{i+1} + \frac{x_i}{2} + 47\right|} - x_i\sin\sqrt{|x_i - (x_{i+1} + 47)|} \right]$ | $\mathbf{x}^* = (512, 404.2319), f(\mathbf{x}^*)! = 959.64$ | $-512 \leq x_i \leq 512$ |
| **CROSS-IN-TRAY MULTIMODAL – MANY LOCAL MINIMA** | $f19(\mathbf{x}) = -0.0001\left(\sin(x_1)\sin(x_2)e^{\left|100 - \left[(x_1^2 + x_2^2)^{0.5}\right]/\pi\right|} + 1\right)^{0.1}$ | $\mathbf{x}^* = (\pm 1.3494066, \pm 1.3494066), f( = 2.0621218$ | $-10 \leq x_i \leq 10$ |
| **GRIEWANK – MULTIMODAL – MANY LOCAL MINIMA** | $f20(\mathbf{x}) = \sum_{i=0}^{D} \frac{x_i^2}{4000} - \prod \cos\left(\frac{x_i}{\sqrt{i}}\right) + 1$ | $\mathbf{x}^* = (0, \dots, 0), f(\mathbf{x}^*) = 0$ | $-100 \leq x_i \leq 100$ |
| **GOLDSTEIN-PRICE MULTIMODAL - UNEQUAL SHAPED** | $f21(\mathbf{x}) = [1 + (x_1 + x_2 + 1)^2(19 - 14x_1 + 3x_1^2 - 14x_2 + \times [30 + (2x_1 - 3x_2)^2(18 - 32x_1 + 12x_1^2 + 48.$ | $\mathbf{x}^* = (0, -1), f(\mathbf{x}^*) = 3$ | $-2 \leq x_i \leq 2$ |
| **DIXON UNIMODAL VALLEY-SHAPED** | $f22(\mathbf{x}) = (x_1 - 1)^2 + \sum_{i=2}^{D} i(2x_i^2 - x_{i-1})^2$ | $\mathbf{x}^* = f(2(2i - 2/2i)), f(\mathbf{x}^*) = 0$ | $-10 \leq x_i \leq 10$ |
| **LEVY – MULTIMODAL – MANY LOCAL MINIMA – CONSTRAINED** | $f23(\mathbf{x}) = \sin^2(\pi w_1) + \sum_{i=1}^{d-1}(w_i - 1)^2[1 + 10\sin^2(\pi w_i + 1)] + (w_d - 1)^2[1 + \sin^2(2\pi w_d)]$ | $\mathbf{x}^* = (1, \dots, 1), f(\mathbf{x}^*) = 0$ | $-10 \leq x_i \leq 10$ |
| **BIRD, MULTIMODAL – CONSTRAINED** | $f24(x) = \sin(x_1)e^{\left[(1 - \cos(x_2))^2\right]} + \cos(x_2)e^{\left[(1 - \sin(x_1))^2\right]} + (x_1 - x_2)^2$ | $\mathbf{x}^* = (4.7, 3.15), f(\mathbf{x}^*) = -106.764537$ | $-2pi \leq x_i \leq 2pi$ |

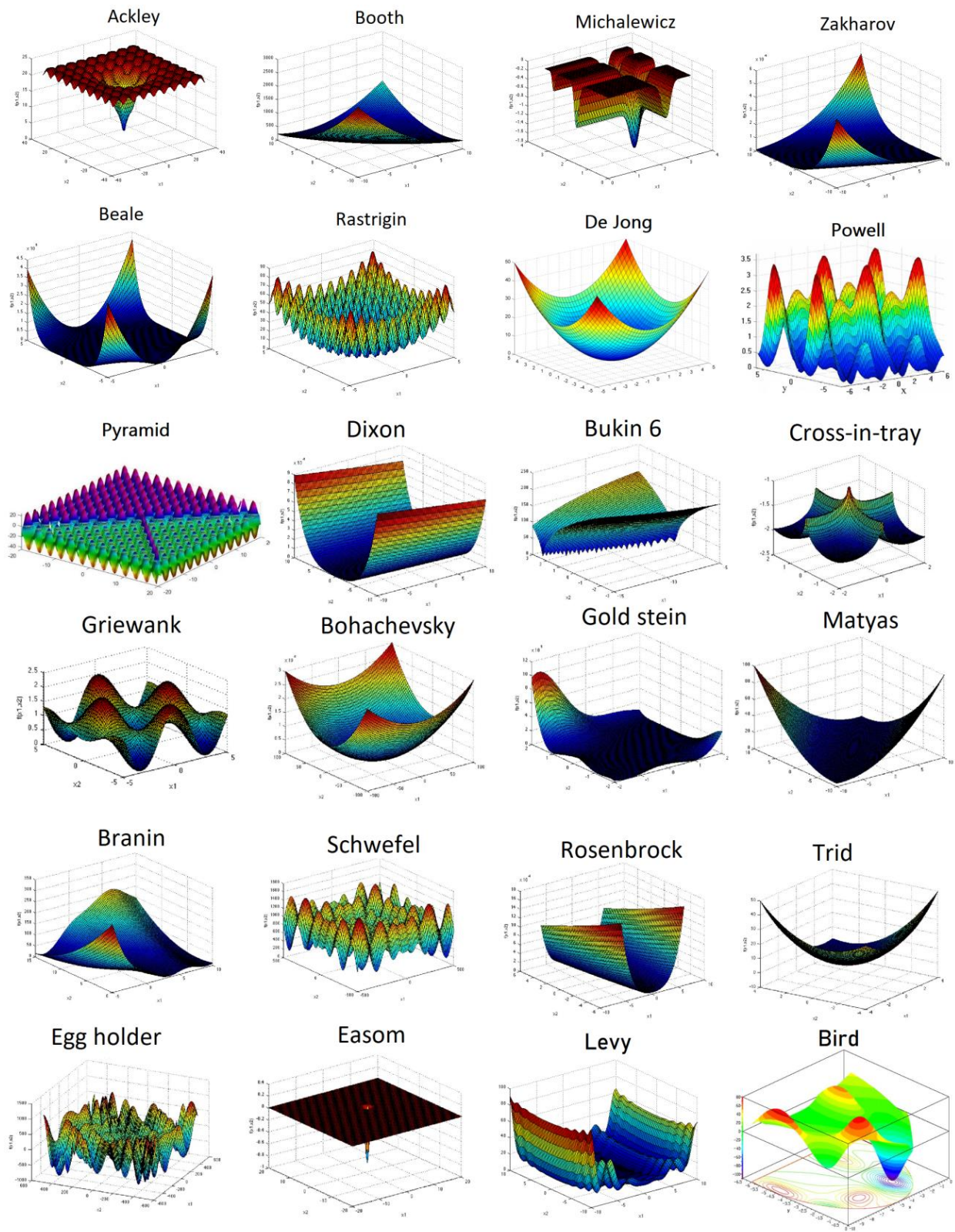

*Figure 4*. 3-D form of the experiment optimization functions for evaluation.

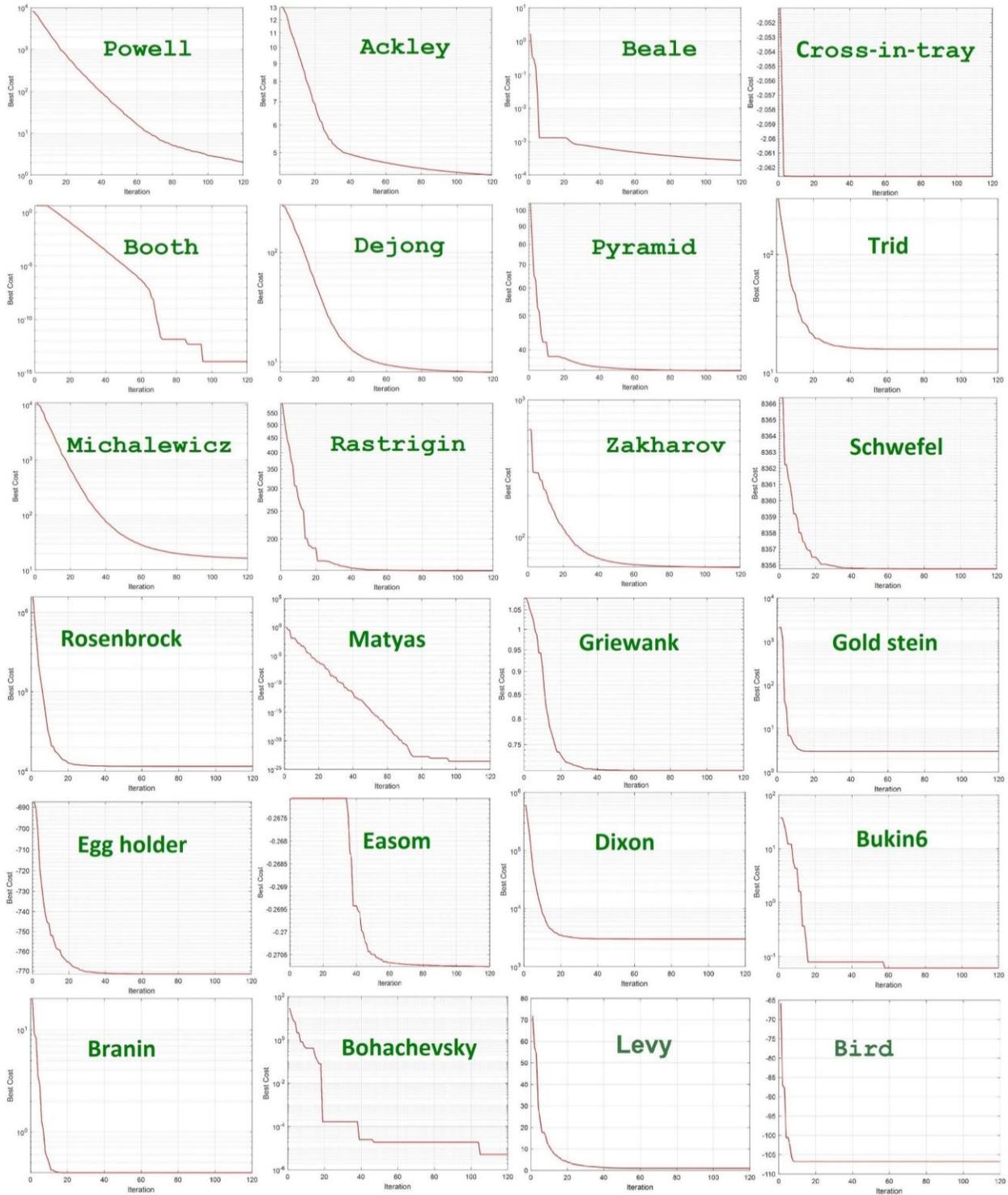

*Figure 5*. VAO algorithm result on nine optimization functions during 120 iterations.

The performance of different algorithms on the Ackley function is presented in Figure6. Figure 7 presents, boxplot of cost functions from results of Figure 6. Now, Figure 8 illustrates performance of VAO algorithm with Rastrigin function with 2, 4, 8, 16, 32 and 64 population over 100 iterations.

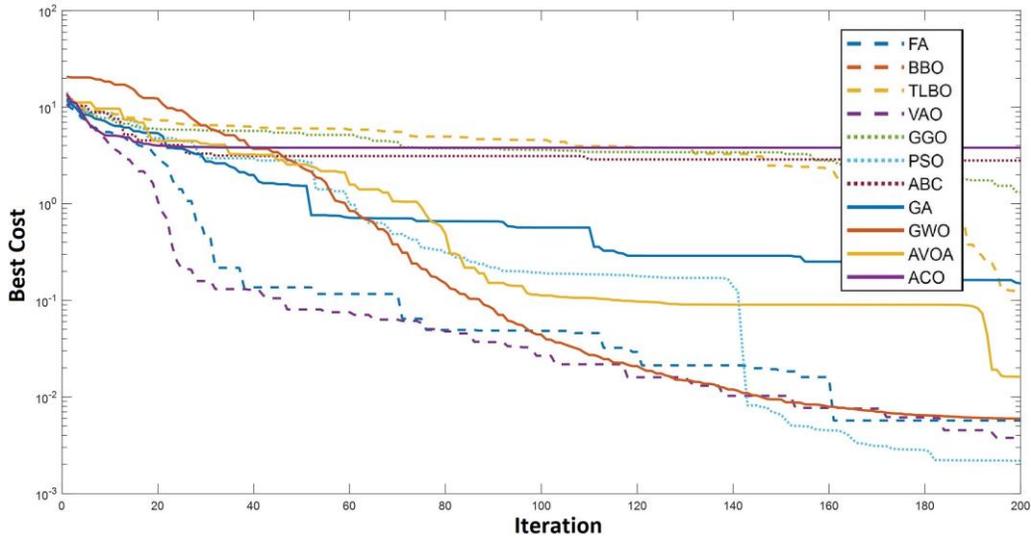

*Figure 6.* Performance comparison of bio-inspired algorithms on the Ackley function.

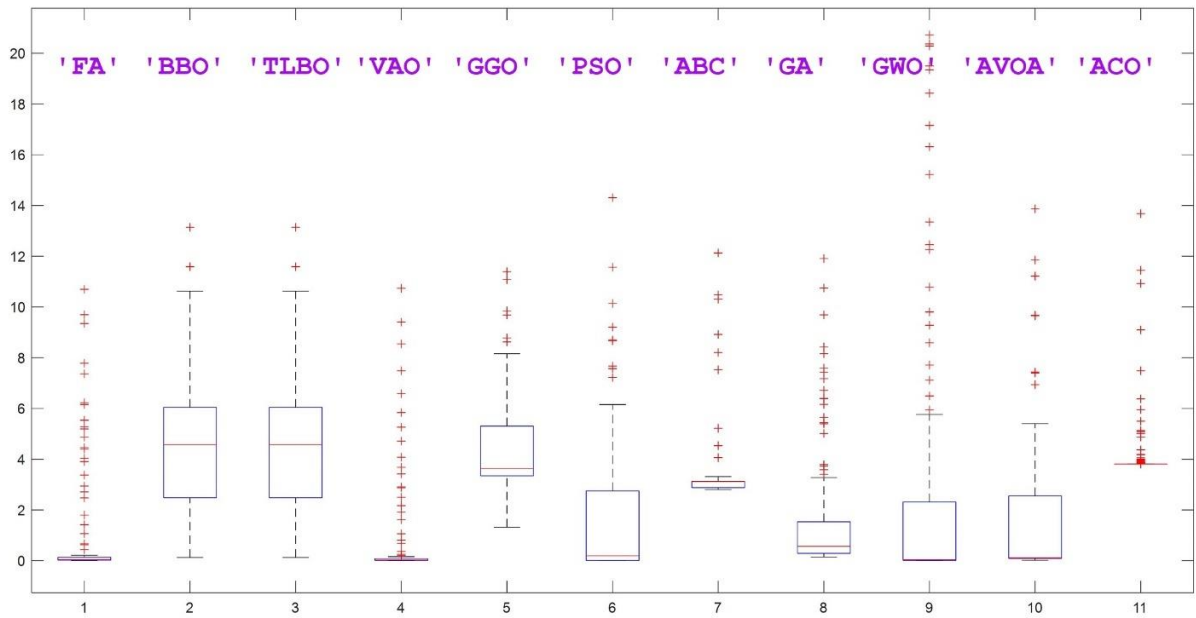

*Figure 7.* boxplot of cost functions from results of Figure 6.

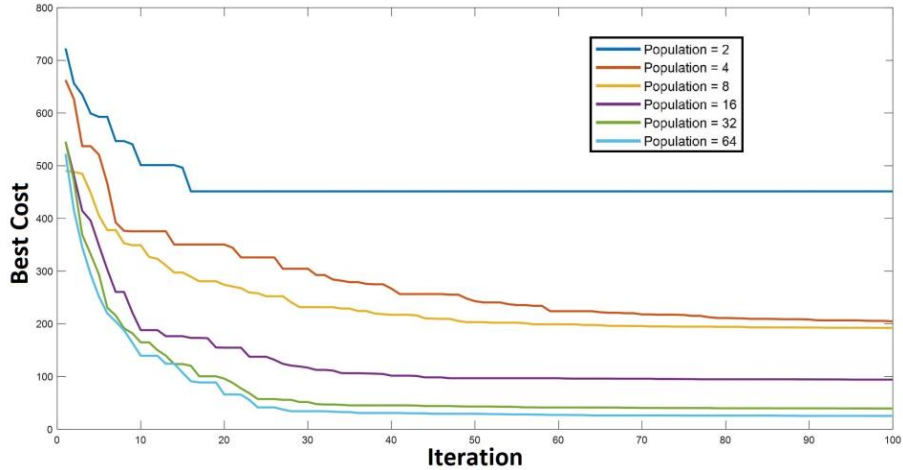

*Figure 8.* VAO algorithm using Rastrigin function with 2, 4, 8, 16, 32 and 64 populations Over 100 iterations.

Figure 9 illustrates the VAO algorithm's performance over 50 generations. The cost values gained after 500 iterations on 24 optimization functions are presented in Table 2 as a comparison for all bio-inspired algorithms. These outcomes are determined by the parameters in Table 3, and the smaller the cost function, the better.

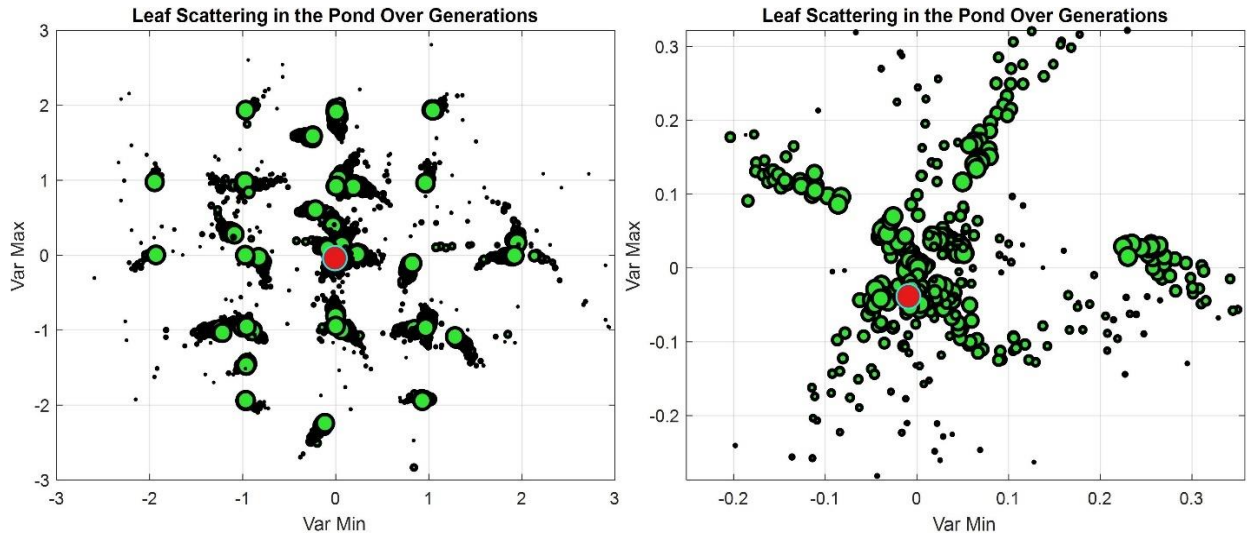

*Figure 9.* Plant scattering using the VAO algorithm over 50 generations in the pond. Left: top view of all generations and alpha plant as red circle. Right: Closer view to the best generation with best cost and alpha plant as red circle. The bigger the plant, the stronger.

Based on Figure 5 results, the performance of the VAO algorithm on all test functions is pretty considerable and promising. On the basis of the values in Table 2, the proposed VAO algorithm could easily compete with the best algorithms. Clearly, PSO, GWO, AVOA, VAO, FA, and GA yielded superior results compared to others. It should be noted that ACO and ABC have the poorest performance. Additionally, TLBO, GGO, and BBO received an average performance ranking overall. Figure 10 depicts the performance of the VAO algorithm with the Ackley and Rastrigin functions. Table 4 presents a run time comparison of different algorithms with the following system specifications: Core i-7 CPU 4 GHz, 16 GB RAM (DDR 3), GTX 1050 Graphic Card (2 GB), and 256 GB SSD Hard Drive.

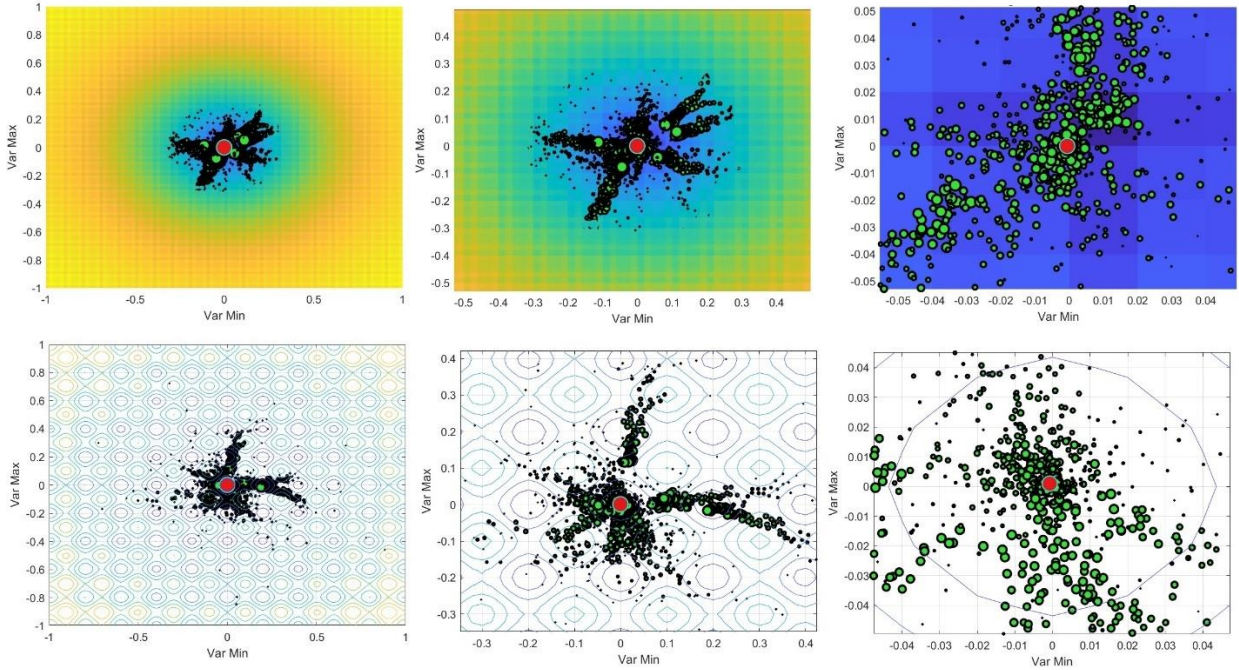

*Figure 10.* Top row: The VAO algorithm on the Ackley function; Bottom row: The VAO algorithm on the Rastrigin function.

Table 2

*Best Cost Value Comparison on All Algorithms after 500 Iterations based on Table 3 Parameters*

| TEST FUNCTION | VALUE | GA | PSO | FA | BBO | ABC | VAO | TLBO | GGO | ACO | GWO | AVOA |
|---|---|---|---|---|---|---|---|---|---|---|---|---|
| 1.ACKLEY | Average | 3.935 | 3.573 | 3.562 | 4.616 | 5.133 | 3.206 | 4.449 | 4.883 | 5.088 | 3.617 | 3.539 |
| | Std | 0.9508 | 0.8377 | 0.8261 | 1.134 | 1.153 | 0.719 | 1.216 | 1.401 | 1.357 | 0.8814 | 0.8032 |
| 2.RASTRIGIN | Average | 167.9 | 153.9 | 153.9 | 195.9 | 223.8 | 139.852 | 182.1 | 196.2 | 210.1 | 144.9 | 138.8 |
| | Std | 31.82 | 29.14 | 29.12 | 37.15 | 42.31 | 26.445 | 34.66 | 37.42 | 39.95 | 25.18 | 29.1 |
| 3.ZAKHAROV | Average | 111.1 | 101.9 | 111.8 | 129.7 | 148.1 | 92.551 | 120.6 | 130.0 | 139.1 | 111.9 | 104.8 |
| | Std | 20.63 | 18.87 | 17.86 | 24.09 | 27.39 | 17.115 | 22.53 | 24.36 | 25.95 | 17.92 | 18.84 |
| 4.BOOTH | Average | 0.09279 | 0.05117 | 0.03964 | 0.1329 | 0.002998 | 0.004 | 0.2864 | 0.4001 | 0.2847 | 0.09493 | 0.01668 |
| | Std | 0.09159 | 0.05007 | 0.03854 | 0.1315 | 0.007798 | 0.003 | 0.2851 | 0.3987 | 0.2832 | 0.09383 | 0.01558 |
| 5.PYRAMID | Average | 11.85 | 10.83 | 9.82 | 13.85 | 15.68 | 9.800 | 13.02 | 14.11 | 14.98 | 10.87 | 13.79 |
| | Std | 1.17 | 1.039 | 1.027 | 1.39 | 1.446 | 0.902 | 1.454 | 1.657 | 1.632 | 1.083 | 1.004 |
| 6.DEJONG | Average | 23.95 | 11.92 | 18.9 | 27.96 | 31.81 | 19.881 | 26.13 | 28.23 | 30.1 | 27.96 | 21.88 |
| | Std | 2.068 | 1.862 | 1.95 | 2.437 | 2.643 | 1.650 | 2.426 | 2.705 | 2.754 | 1.906 | 1.827 |
| 7.BEALE | Average | 0.08799 | 0.04677 | 0.03524 | 0.1273 | 0.002998 | 0.000 | 0.2812 | 0.3945 | 0.2787 | 0.09053 | 0.01228 |
| | Std | 0.09039 | 0.04897 | 0.03744 | 0.1301 | 0.006198 | 0.002 | 0.2838 | 0.3973 | 0.2817 | 0.09273 | 0.01448 |
| 8.POWELL | Average | 188.4 | 172.6 | 166.6 | 219.8 | 251.0 | 156.889 | 204.2 | 220.0 | 235.6 | 162.7 | 179.6 |
| | Std | 28.19 | 25.81 | 25.2 | 32.91 | 37.47 | 23.418 | 30.72 | 33.18 | 35.41 | 25.85 | 25.77 |
| 9.MICHALEWICZ | Average | -6.112 | -8.387 | -9.398 | -5.61 | -2.26 | -7.667 | -5.686 | -3.34 | -3.22 | -9.343 | -8.421 |
| | Std | 2.742 | 1.48 | 2.168 | 3.224 | 3.542 | 2.212 | 3.157 | 3.491 | 3.597 | 2.524 | 2.445 |
| 10.MATYAS | Average | 0.1444 | 0.09847 | 0.08694 | 0.1931 | 0.0782 | 0.047 | 0.3423 | 0.4603 | 0.3492 | 0.1422 | 0.06398 |
| | Std | 0.1276 | 0.08307 | 0.07154 | 0.1735 | 0.0558 | 0.033 | 0.3241 | 0.4407 | 0.3282 | 0.1268 | 0.04858 |
| 11.TRID | Average | -53.83 | -54.88 | -57.89 | -50.77 | -17.89 | -58.930 | -50.63 | -40.51 | -25.62 | -52.83 | -51.91 |
| | Std | 8.333 | 7.605 | 7.593 | 9.747 | 11.0 | 6.871 | 9.214 | 10.01 | 10.59 | 7.649 | 7.57 |
| 12.SCHWEFEL | Average | 9996.0 | 9262.0 | 9163.0 | 11780.0 | 13330.0 | 8330.001 | 10830.0 | 11660.0 | 12500.0 | 8163.0 | 9111.0 |
| | Std | 182.5 | 167.3 | 164.2 | 212.9 | 243.2 | 152.011 | 197.9 | 213.2 | 228.3 | 155.3 | 177.2 |
| 13.EASOM | Average | -1.412 | -1.328 | -1.34 | -1.623 | -0.997 | -1.450 | -0.344 | -1.955 | -0.596 | -1.284 | -1.363 |

| | | GA | PSO | FA | BBO | ABC | VAO | TLBO | GGO | ACO | GWO | AVOA |
|---|---|---|---|---|---|---|---|---|---|---|---|---|
| | Std | 0.328 | 0.2668 | 0.2552 | 0.4073 | 0.323 | 0.200 | 0.5412 | 0.6745 | 0.5787 | 0.3105 | 0.2323 |
| **14.ROSENBROCK** | Average | 0.08919 | 0.04787 | 0.03634 | 0.1287 | 0.004598 | 0.001 | 0.2825 | 0.3959 | 0.2802 | 0.09163 | 0.01338 |
| | Std | 0.09519 | 0.05337 | 0.04184 | 0.1357 | 0.0126 | 0.006 | 0.289 | 0.4029 | 0.2877 | 0.09713 | 0.01888 |
| **15.BOHACHEVSKY** | Average | 1.245 | 1.107 | 1.096 | 1.477 | 1.545 | 0.964 | 1.534 | 1.744 | 1.725 | 1.151 | 1.073 |
| | Std | 0.4744 | 0.401 | 0.3894 | 0.5781 | 0.5182 | 0.322 | 0.6998 | 0.8453 | 0.7617 | 0.4447 | 0.3665 |
| **16.BUKIN 6** | Average | 0.1372 | 0.09187 | 0.08034 | 0.1847 | 0.0686 | 0.041 | 0.3345 | 0.4519 | 0.3402 | 0.1356 | 0.05738 |
| | Std | 0.1312 | 0.08637 | 0.07484 | 0.1777 | 0.0606 | 0.036 | 0.328 | 0.4449 | 0.3327 | 0.1301 | 0.05188 |
| **17.BRANIN** | Average | 0.2776 | 0.2206 | 0.209 | 0.3485 | 0.2558 | 0.158 | 0.4866 | 0.6157 | 0.5157 | 0.2643 | 0.1861 |
| | Std | 0.208 | 0.1568 | 0.1452 | 0.2673 | 0.163 | 0.100 | 0.4112 | 0.5345 | 0.4287 | 0.2005 | 0.1223 |
| **18.EGG HOLDER** | Average | -772.0 | -891.1 | -809.1 | -750.0 | -221.0 | -810.112 | -605.0 | -800.0 | -352.0 | -831.0 | -822.1 |
| | Std | 60.09 | 55.05 | 52.04 | 70.13 | 80.01 | 50.004 | 65.29 | 70.4 | 75.28 | 50.09 | 51.02 |
| **19.CROSS-IN-TRAY** | Average | -2.077 | -2.213 | -2.224 | -1.748 | -0.283 | -2.054 | -1.389 | -1.481 | -0.802 | -2.169 | -2.247 |
| | Std | 0.784 | 0.6848 | 0.6732 | 0.9393 | 0.931 | 0.580 | 1.035 | 1.207 | 1.149 | 0.7285 | 0.6503 |
| **20.GRIEWANK** | Average | 0.7276 | 0.6331 | 0.6215 | 0.8735 | 0.8558 | 0.533 | 0.9741 | 1.141 | 1.078 | 0.6768 | 0.5986 |
| | Std | 0.2932 | 0.2349 | 0.2233 | 0.3667 | 0.2766 | 0.171 | 0.5035 | 0.6339 | 0.5352 | 0.2786 | 0.2004 |
| **21.GOLDSTEIN** | Average | 0.1168 | 0.07317 | 0.06164 | 0.1609 | 0.0414 | 0.024 | 0.3124 | 0.4281 | 0.3147 | 0.1169 | 0.03868 |
| | Std | 0.2344 | 0.181 | 0.1694 | 0.2981 | 0.1982 | 0.122 | 0.4398 | 0.5653 | 0.4617 | 0.2247 | 0.1465 |
| **22.DIXON** | Average | 0.148 | 0.1018 | 0.09024 | 0.1973 | 0.083 | 0.050 | 0.3462 | 0.4645 | 0.3537 | 0.1455 | 0.06728 |
| | Std | 0.1024 | 0.05997 | 0.04844 | 0.1422 | 0.0222 | 0.012 | 0.2968 | 0.4113 | 0.2967 | 0.1037 | 0.02548 |
| **2.LEVY** | Average | 1.981 | 1.099 | 1.411 | 1.000 | 2.662 | 0.024 | 1.443 | 2.053 | 2.025 | 0.601 | 0.520 |
| | Std | 0.802 | 0.052 | 0.811 | 0.035 | 1.012 | 0.012 | 0.313 | 0.998 | 1.000 | 0.099 | 0.011 |
| **24.BIRD** | Average | -90.541 | -102.660 | -105.882 | -104.220 | -96.145 | -107.117 | -103.991 | -99.885 | -90.365 | -106.422 | -107.101 |
| | Std | 9.500 | 8.441 | 5.517 | 6.857 | 20.255 | 3.250 | 5.990 | 8.696 | 15.332 | 4.170 | -4.600 |

Table 3

*Algorithms' Parameters for the Main Experiment*

| PARAMETERS | GA | PSO | FA | BBO | ABC | VAO | TLBO | GGO | ACO | GWO | AVOA |
|---|---|---|---|---|---|---|---|---|---|---|---|
| DECISION VARIABLES (DV) | 15 | 15 | 15 | 15 | 15 | 15 | 15 | 15 | 15 | 15 | 15 |
| DECISION VARIABLES SIZE | [1,15] | [1,15] | [1,15] | [1,15] | [1,15] | [1,15] | [1,15] | [1,15] | [1,15] | [1,15] | [1,15] |
| LOWER BOUND OF VARIABLES (LV) | -10 | -10 | -10 | -10 | -10 | -10 | -10 | -10 | -10 | -10 | -10 |
| UPPER BOUND OF VARIABLES (UP) | 10 | 10 | 10 | 10 | 10 | 10 | 10 | 10 | 10 | 10 | 10 |
| ITERATIONS | 500 | 500 | 500 | 500 | 500 | 500 | 500 | 500 | 500 | 500 | 500 |
| POPULATION SIZE (P) | 20 | 20 | 20 | 20 | 20 | 20 | 20 | 20 | 20 | 20 | 20 |
| CROSSOVER PERCENTAGE (PC) | 0.7 | - | - | - | - | - | - | - | - | - | - |
| NUMBER OF OFFSPRING'S (PARENTS) | 2*(PC*P/2) | - | - | - | - | - | - | - | - | - | - |
| MUTATION PERCENTAGE (MP) | 0.3 | - | - | - | - | - | - | - | 0.3 | - | - |
| NUMBER OF MUTANTS | MP*P | - | - | - | - | - | - | - | - | - | - |
| MUTATION RATE | 0.2 | 0.2 | 0.2 | 0.2 | 0.2 | 0.2 | 0.2 | 0.2 | 0.2 | 0.2 | 0.2 |
| INERTIA WEIGHT | - | 1 | - | - | - | - | - | - | - | - | - |
| INERTIA WEIGHT DAMPING RATIO | - | 0.99 | - | - | - | - | - | - | - | - | - |
| PERSONAL LEARNING COEFFICIENT | - | 1.5 | - | - | - | - | - | - | - | - | - |
| GLOBAL LEARNING COEFFICIENT | - | 2 | - | - | - | - | - | - | - | - | - |
| LIGHT ABSORPTION COEFFICIENT | - | - | 1 | - | - | - | - | - | - | - | - |
| ATTRACTION COEFFICIENT | - | - | 2 | - | - | - | - | - | - | - | - |
| MUTATION DAMPING RATIO | - | - | 0.98 | - | - | 0.99 | - | - | - | - | - |
| KEEP RATE (KR) | - | - | - | 0.2 | - | - | - | - | - | - | - |
| KEPT HABITATS | - | - | - | KR*P | - | - | - | - | - | - | - |
| NEW HABITATS | - | - | - | P-KR | - | - | - | - | - | - | - |
| EMMIGRATION RATES (ER) | - | - | - | 0.2 | - | - | - | - | - | - | - |
| IMMIGRATION RATES | - | - | - | 1-(ER) | - | - | - | - | - | - | - |
| ONLOOKER BEES | - | - | - | - | P | - | - | - | - | - | - |
| ABANDONMENT LIMIT PARAMETER | - | - | - | - | 0.6*DV*P | - | - | - | - | - | - |
| ACCELERATION COEFFICIENT UPPER | - | - | - | - | 1 | - | - | - | - | - | - |
| DRAWBACK COEFFICIENT 1 | - | - | - | - | - | [1 3] | - | - | - | - | - |
| DRAWBACK COEFFICIENT 2 | - | - | - | - | - | [1 3] | - | - | - | - | - |
| INTRA COMPETITION RATE | - | - | - | - | - | [10 30] | - | - | - | - | - |
| NUMBER OF ROCKS (NR) | - | - | - | - | - | - | - | 500 | - | - | - |
| NUMBER OF RUBBLES | - | - | - | - | - | - | - | P-NR | - | - | - |
| SELECTION PRESSURE | - | - | - | - | - | - | - | 1 | 0.5 | - | - |

Table 4

*Run time comparison results for all algorithms on all optimization functions (in seconds)*

| | | GA | PSO | FA | BBO | ABC | VAO | TLBO | GGO | ACO | GWO | AVOA |
|---|---|---|---|---|---|---|---|---|---|---|---|---|
| **ACKLEY** | Avg | 1.012 | 0.750 | 3.521 | 0.882 | 1.315 | 1.254 | 0.784 | 1.933 | 1.692 | 0.581 | 0.463 |
| | Std | 0.175 | 0.024 | 1.020 | 0.080 | 0.412 | 0.115 | 0.139 | 0.360 | 0.251 | 0.113 | 0.170 |
| **RASTRIGIN** | Avg | 0.973 | 0.628 | 3.108 | 0.623 | 1.100 | 0.960 | 0.514 | 1.721 | 1.555 | 0.390 | 0.400 |
| | Std | 0.097 | 0.066 | 0.690 | 0.075 | 0.214 | 0.079 | 0.103 | 0.184 | 0.197 | 0.087 | 0.096 |
| **ZAKHAROV** | Avg | 0.931 | 0.700 | 3.447 | 0.799 | 1.218 | 1.004 | 0.520 | 1.817 | 1.602 | 0.412 | 0.580 |
| | Std | 0.099 | 0.053 | 0.740 | 0.095 | 0.558 | 0.088 | 0.091 | 0.298 | 0.610 | 0.091 | 0.077 |
| **BOOTH** | Avg | 0.869 | 0.512 | 3.009 | 0.603 | 1.117 | 0.881 | 0.440 | 1.610 | 1.497 | 0.400 | 0.517 |
| | Std | 0.139 | 0.079 | 0.373 | 0.094 | 0.271 | 0.144 | 0.101 | 0.269 | 0.401 | 0.092 | 0.098 |
| **PYRAMID** | Avg | 0.857 | 0.509 | 3.115 | 0.614 | 1.121 | 0.899 | 0.591 | 1.888 | 1.471 | 0.580 | 0.607 |
| | Std | 0.097 | 0.088 | 0.229 | 0.081 | 0.227 | 0.085 | 0.065 | 0.231 | 0.118 | 0.079 | 0.071 |
| **DEJONG** | Avg | 0.841 | 0.528 | 3.100 | 0.771 | 1.350 | 1.119 | 0.536 | 1.912 | 1.670 | 0.650 | 0.709 |
| | Std | 0.108 | 0.096 | 0.413 | 0.087 | 0.288 | 0.089 | 0.097 | 0.254 | 0.266 | 0.081 | 0.085 |
| **BEALE** | Avg | 0.665 | 0.470 | 2.411 | 0.591 | 1.008 | 0.764 | 0.388 | 1.152 | 1.207 | 0.254 | 0.299 |
| | Std | 0.038 | 0.049 | 0.511 | 0.047 | 0.111 | 0.060 | 0.055 | 0.217 | 0.213 | 0.036 | 0.039 |
| **POWELL** | Avg | 0.755 | 0.703 | 2.909 | 0.870 | 1.250 | 1.120 | 0.590 | 1.359 | 1.405 | 0.578 | 0.608 |
| | Std | 0.083 | 0.081 | 0.254 | 0.089 | 0.100 | 0.165 | 0.071 | 0.102 | 0.109 | 0.097 | 0.091 |
| **MICHALEWICZ** | Avg | 0.997 | 0.933 | 3.257 | 1.087 | 1.444 | 1.139 | 0.588 | 1.641 | 1.629 | 0.709 | 0.857 |
| | Std | 0.081 | 0.076 | 0.270 | 0.101 | 0.133 | 0.076 | 0.061 | 0.138 | 0.127 | 0.052 | 0.049 |
| **MATYAS** | Avg | 1.100 | 0.902 | 2.278 | 1.118 | 1.499 | 1.190 | 0.618 | 2.008 | 1.755 | 0.698 | 0.737 |
| | Std | 0.088 | 0.075 | 0.120 | 0.118 | 0.136 | 0.071 | 0.069 | 0.137 | 0.148 | 0.067 | 0.060 |
| **TRID** | Avg | 1.001 | 0.833 | 2.000 | 1.114 | 1.290 | 0.826 | 0.543 | 1.831 | 1.550 | 0.420 | 0.611 |
| | Std | 0.087 | 0.069 | 0.114 | 0.100 | 0.108 | 0.055 | 0.041 | 0.165 | 0.199 | 0.032 | 0.035 |
| **SCHWEFEL** | Avg | 1.210 | 1.119 | 2.449 | 1.320 | 1.561 | 1.009 | 0.812 | 2.150 | 1.880 | 0.664 | 0.800 |
| | Std | 0.059 | 0.053 | 0.177 | 0.099 | 0.98 | 0.067 | 0.051 | 0.161 | 0.102 | 0.060 | 0.049 |
| **EASOM** | Avg | 0.899 | 1.000 | 2.285 | 1.117 | 1.351 | 0.700 | 0.419 | 1.751 | 1.310 | 0.402 | 0.609 |
| | Std | 0.024 | 0.036 | 0.070 | 0.077 | 0.073 | 0.023 | 0.031 | 0.057 | 0.062 | 0.029 | 0.038 |
| **ROSENBROCK** | Avg | 1.228 | 1.304 | 2.533 | 1.335 | 1.695 | 0.950 | 0.630 | 1.957 | 1.444 | 0.600 | 0.618 |
| | Std | 0.098 | 0.090 | 0.255 | 0.281 | 0.200 | 0.091 | 0.079 | 0.199 | 0.173 | 0.039 | 0.052 |
| **BOHACHEVSKY** | Avg | 1.114 | 1.227 | 2.337 | 1.186 | 1.550 | 0.699 | 0.489 | 1.677 | 1.245 | 0.459 | 0.471 |
| | Std | 0.080 | 0.083 | 0.366 | 0.301 | 0.309 | 0.041 | 0.055 | 0.211 | 0.206 | 0.044 | 0.047 |
| **BUKIN 6** | Avg | 1.218 | 1.211 | 2.229 | 1.174 | 1.599 | 0.657 | 0.499 | 1.648 | 1.239 | 0.522 | 0.574 |
| | Std | 0.88 | 0.086 | 0.099 | 0.091 | 0.097 | 0.069 | 0.060 | 0.109 | 0.152 | 0.080 | 0.029 |
| **BRANIN** | Avg | 1.475 | 1.561 | 2.891 | 1.225 | 1.878 | 0.863 | 0.600 | 1.999 | 1.561 | 0.661 | 0.599 |
| | Std | 0.075 | 0.068 | 0.092 | 0.091 | 0.087 | 0.066 | 0.054 | 0.084 | 0.085 | 0.065 | 0.036 |
| **EGG HOLDER** | Avg | 1.711 | 1.788 | 3.722 | 1.653 | 2.225 | 1.447 | 0.957 | 2.578 | 1.964 | 0.994 | 0.896 |
| | Std | 0.106 | 0.097 | 0.548 | 0.369 | 0.214 | 0.099 | 0.073 | 0.255 | 0.200 | 0.076 | 0.074 |
| **CROSS-IN-TRAY** | Avg | 1.330 | 1.497 | 2.955 | 1.344 | 1.833 | 0.997 | 0.581 | 1.930 | 1.412 | 0.633 | 0.673 |
| | Std | 0.100 | 0.093 | 0.299 | 0.246 | 0.201 | 0.057 | 0.066 | 0.203 | 0.209 | 0.051 | 0.057 |
| **GRIEWANK** | Avg | 1.108 | 1.203 | 2.500 | 1.114 | 1.777 | 0.882 | 0.513 | 1.881 | 1.369 | 0.594 | 0.670 |
| | Std | 0.090 | 0.096 | 0.130 | 0.108 | 0.102 | 0.081 | 0.083 | 0.101 | 0.105 | 0.076 | 0.099 |
| **GOLDSTEIN** | Avg | 1.003 | 1.191 | 2.436 | 1.005 | 1.510 | 0.901 | 0.491 | 1.714 | 1.267 | 0.408 | 0.443 |
| | Std | 0.063 | 0.060 | 0.084 | 0.079 | 0.095 | 0.034 | 0.070 | 0.090 | 0.095 | 0.059 | 0.067 |
| **DIXON** | Avg | 1.203 | 1.211 | 2.779 | 1.203 | 1.882 | 1.087 | 0.566 | 1.886 | 1.299 | 0.477 | 0.482 |
| | Std | 0.066 | 0.043 | 0.110 | 0.125 | 0.101 | 0.068 | 0.042 | 0.114 | 0.086 | 0.035 | 0.039 |
| **LEVY** | Avg | 1.182 | 1.025 | 2.542 | 1.369 | 1.587 | 1.055 | 0.605 | 1.652 | 1.555 | 0.694 | 0.725 |
| | Std | 0.056 | 0.091 | 0.210 | 0.129 | 0.146 | 0.087 | 0.078 | 0.129 | 0.133 | 0.063 | 0.060 |
| **BIRD** | Avg | 1.024 | 1.099 | 2.879 | 2.150 | 1.860 | 1.102 | 0.607 | 1.546 | 1.473 | 0.552 | 0.591 |
| | Std | 0.052 | 0.075 | 0.091 | 0.090 | 0.080 | 0.064 | 0.087 | 0.086 | 0.099 | 0.058 | 0.041 |

In order to further validate the algorithm, the VAO algorithm is tested with 10 optimization problems in different domains, namely Minimum Spanning Tree (MST) (Lu *et al*., 2022), Hub Location Allocation (HLA) (Ghaffarinasab & Kara, 2022), Quadratic Assignment Problem (QAP) (Koopmans & Beckmann, 1957), (Guo *et al*., 2020), Clustering (Mousavi *et al*., 2017b), Feature selection (Mahalakshmi, D., *et al*, 2022), Regression (Barkhordar, Zahra, *et al*, 2022), Economic Dispatching (ED) (Yuan et al., 2019), Parallel Machine Scheduling (PMS) (Cheng et al., 1990), Image Quantization (IQ) (Orchard et al., 1991), and Image Segmentation (IS) (Cheng et al., 2001) and compared with PSO, FA, DE, HS, and ICA algorithms on the same data.

Due to the various behavior of metaheuristic algorithms, it is decided to select at least one algorithm from each category to compare our algorithm with it. However, those selected algorithms should be famous and benchmarked, and selected carefully to address this issue. Metaheuristics are inspired by the nature and it categorized into four main categories of evolutionary-based, human-based, swarm-based, and physique-based algorithms. FA and PSO are selected from the swarm-based category and the proposed VAO is from the same category. DE is selected from the evolutionary-based category, HS is selected from the physique-based category and ICA is selected from the human-based category. All mentioned comparing algorithms are experimented on all 10 problems and could solve all problems but with different qualities as below.

Minimum spanning Tree (MST) (Lu *et al*., 2022) is a well-known problem in graph theory that entails locating a connected path between numerous vertices with the lowest edge weight and no vertex cycles. This optimization problem could be tackled by standard algorithms such as Prim's or by a nature-inspired algorithm that iteratively optimizes the cost function. Solving graph theory problems like MST by intelligent optimization algorithms returns much better results as degree-constrained MST has NP-hard complexity. Figure 11 depicts the comparison of the proposed VAO for MST with PSO for MST (Guo *et al.*, 2013), FA for MST (Lin *et al*., 2020) DE for MST (Sandeep et al., 2020), HS for MST (Wu et al., 2021), and ICA for MST (Hosseini et al., 2012) on 22 vertices in the range of 0 to 100. All algorithms executed over 200 iterations with an initial population of 35 individuals. Looking at the figure, PSO, FA, and VAO could solve the MST problem at a relatively low cost but DE, HS, and ICA failed to solve the problem. However, DE just failed to solve one vertex (20, 20), but ICA showed the worst performance with the highest cost value among all algorithms. Additionally, VAO achieved the lowest cost value but, the highest runtime. Table 5 presents comparison results of all algorithms on MST problem.

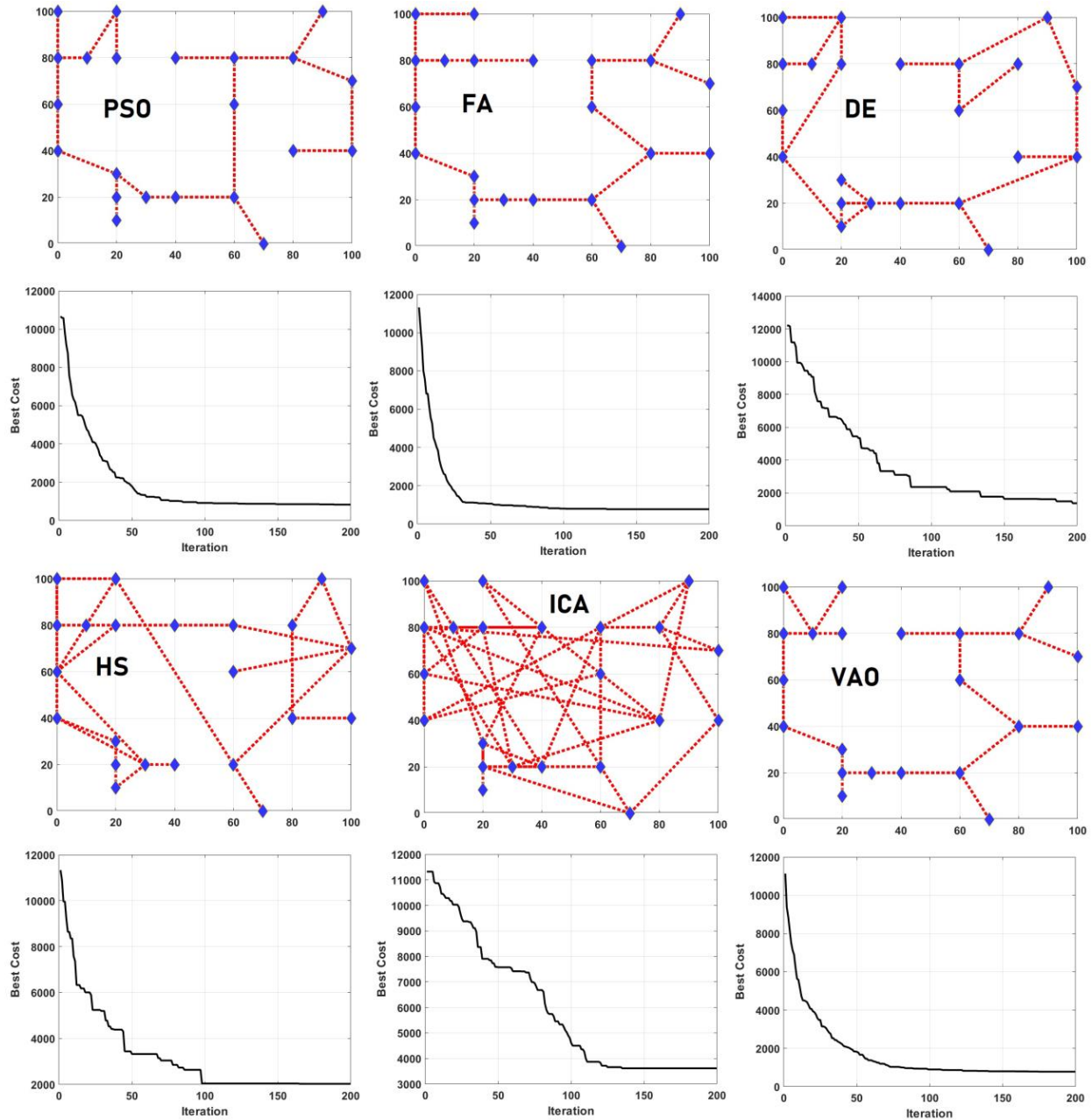

*Figure 11.* FA, PSO, DE, HS, ICA, and VAO algorithm solutions on the same data for the MST optimization problem (200 iterations and 35 population).

Table 5

*Comparison results of all algorithms on MST problem with same parameters*

| | | PSO | FA | DE | HS | ICA | VAO |
|---|---|---|---|---|---|---|---|
| **IS PROBLEM SOLVED?** | | ✓ | ✓ | ✕ | ✕ | ✕ | ✓ |
| **COST VALUE** | Avg | 865 | 763 | 1292 | 2050 | 3865 | 748 |
| | Std | 35 | 29 | 154 | 256 | 329 | 11 |
| **RUN TIME** | Avg | 3.932 (s) | 7.048 (s) | 1.445 (s) | 3.620 (s) | 1.362 (s) | 9.360 (s) |
| | Std | 0.27 | 1.24 | 0.19 | 0.58 | 0.20 | 2.02 |

One of the booming study fields in location theory is Hub Location Allocation (HLA) (Ghaffarinasab & Kara, 2022), which is an NP-Hard problem that cannot be handled with conventional methods and must be tackled using intelligent algorithms for big sizes. The problem could be presented simply in a facilities-client relationship. Multiple points or clients are considered in a two-dimensional picture, and these clients should get services from their facilities, such as Internet Service Provider (ISP) facilities. The solution is to find the shortest route between clients and facilities so that all clients receive services from a small number of facilities. Here, 40 vertices (black triangles) or points are scattered, and algorithms determine how to find the facilities (green circles) and their associated clients. Figure 12 compares the proposed VAO for HLA to the PSO for HLA (Saeidian *et al.*, 2018), FA for HLA (Li *et al.*, 2019), DE for HLA (Atta et al., 2020), HS for HLA (Hajipour et al., 2014), and ICA for HLA (Mohammadi et al.,2011). Six solutions underwent 200 iterations with a population value of 40 individuals. Also, Table 6 represents comparison results for the HLA problem by all algorithms. By looking at Figure 12, just DE and ICA could not solve the problem as their behavior leads to finding the shortest path for all clients. However, just like MST problem, the proposed VAO algorithm achieved the lowest cost value but, the highest run time. It has to be mentioned that HS could solve the problem but, borderline.

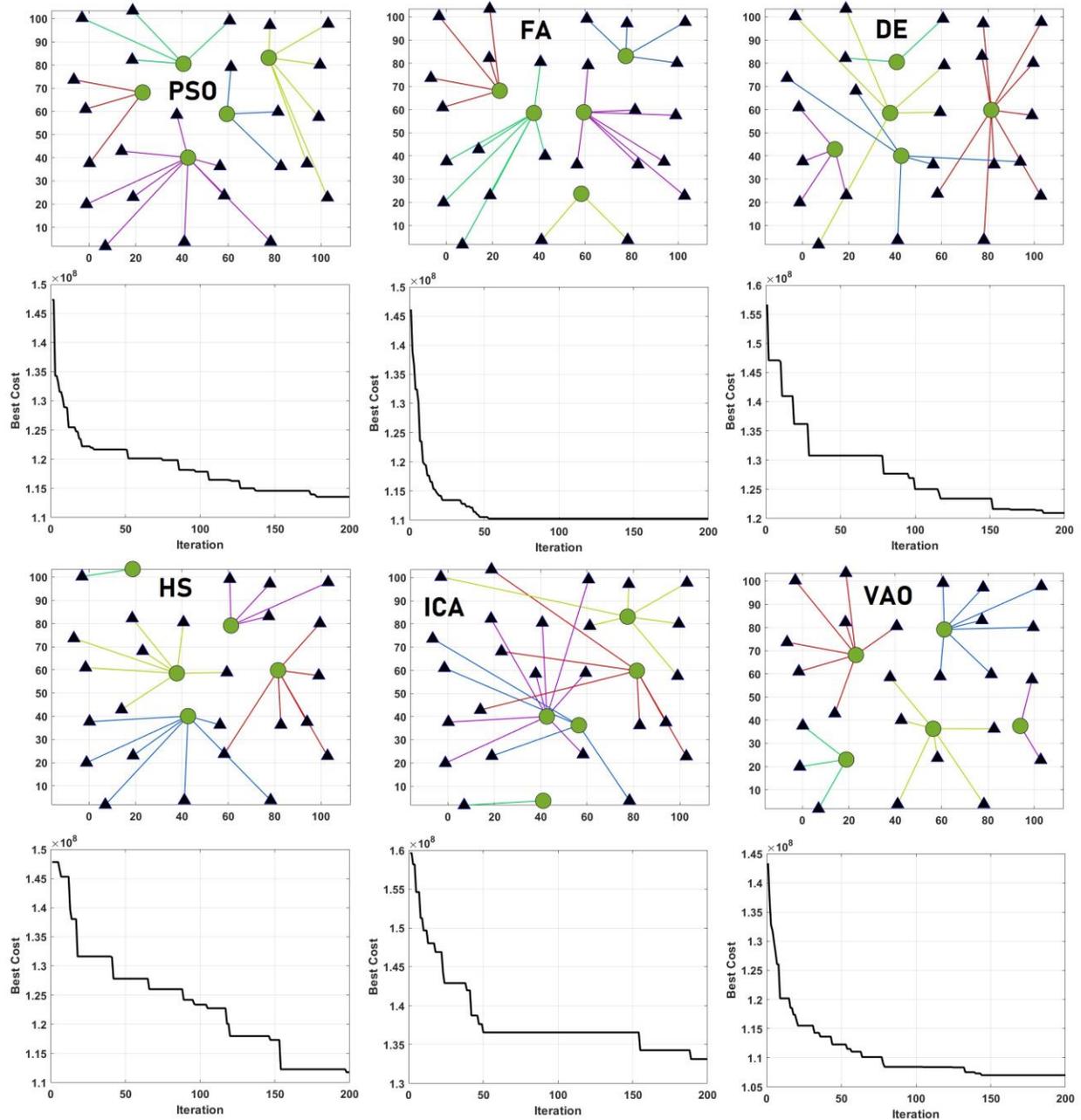

*Figure 12.* FA, PSO, DE, HS, ICA and VAO algorithm solutions on the same data for the HLA optimization problem (200 iteration and 40 population).

Table 6

*Comparison results of all algorithms on HLA problem with same parameters*

|  |  | PSO | FA | DE | HS | ICA | VAO |
|---|---|---|---|---|---|---|---|
| **IS PROBLEM SOLVED?** |  | ✓ | ✓ | ✗ | ✓ | ✗ | ✓ |
| **COST VALUE** | Avg | 1.09e+08 | 1.10e+08 | 1.20e+08 | 1.11e+08 | 1.33e+08 | 1.07e+08 |
|  | Std | 9.43e+06 | 9.02e+06 | 9.70e+06 | 9.23e+06 | 9.15e+06 | 9.63e+06 |
| **RUN TIME** | Avg | 12.479 (s) | 26.033 (s) | 6.781 (s) | 9.868 (s) | 5.821 (s) | 35.111 (s) |
|  | Std | 1.52 | 3.62 | 0.87 | 0.91 | 0.73 | 2.39 |

Similar to HLA, Quadratic Assignment Problem (QAP) is a combinational optimization problem in operational research in mathematics introduced by a facility location problem (Koopmans & Beckmann, 1957). Considering n locations and m facilities, there exists a distance value and a weight for each pair of locations and facilities, respectively. The objective is to assign all amenities to locations while minimizing the sum of distances multiplied by the relevant weight. Here, we scattered 40 samples, and algorithms divided them by location and facility. Figure 13 depicts some solutions to the same problem generated by PSO for QAP (Mamaghani & Meybodi, 2012), FA for QAP (Guo *et al*., 2020), DE for QAP (Kushida et al., 2012), HS for QAP (Salman et al., 2013), ICA for QAP (Hosseini et al., 2014), and VOA for QAP. The population for every algorithm was 50, with over 400 iterations. Table 7 holds comparison results for the QAP problem by all algorithms. Algorithms DE, HS, and ICA failed to solve the problem but, PSO, FA, and VAO succeeded to solve the same problem. FA and VAO could converge at the beginning iterations, however, VAO could achieve better cost value than FA. Also, ICA showed the worse performance, HS was the fastest algorithm, and VAO the slowest one.

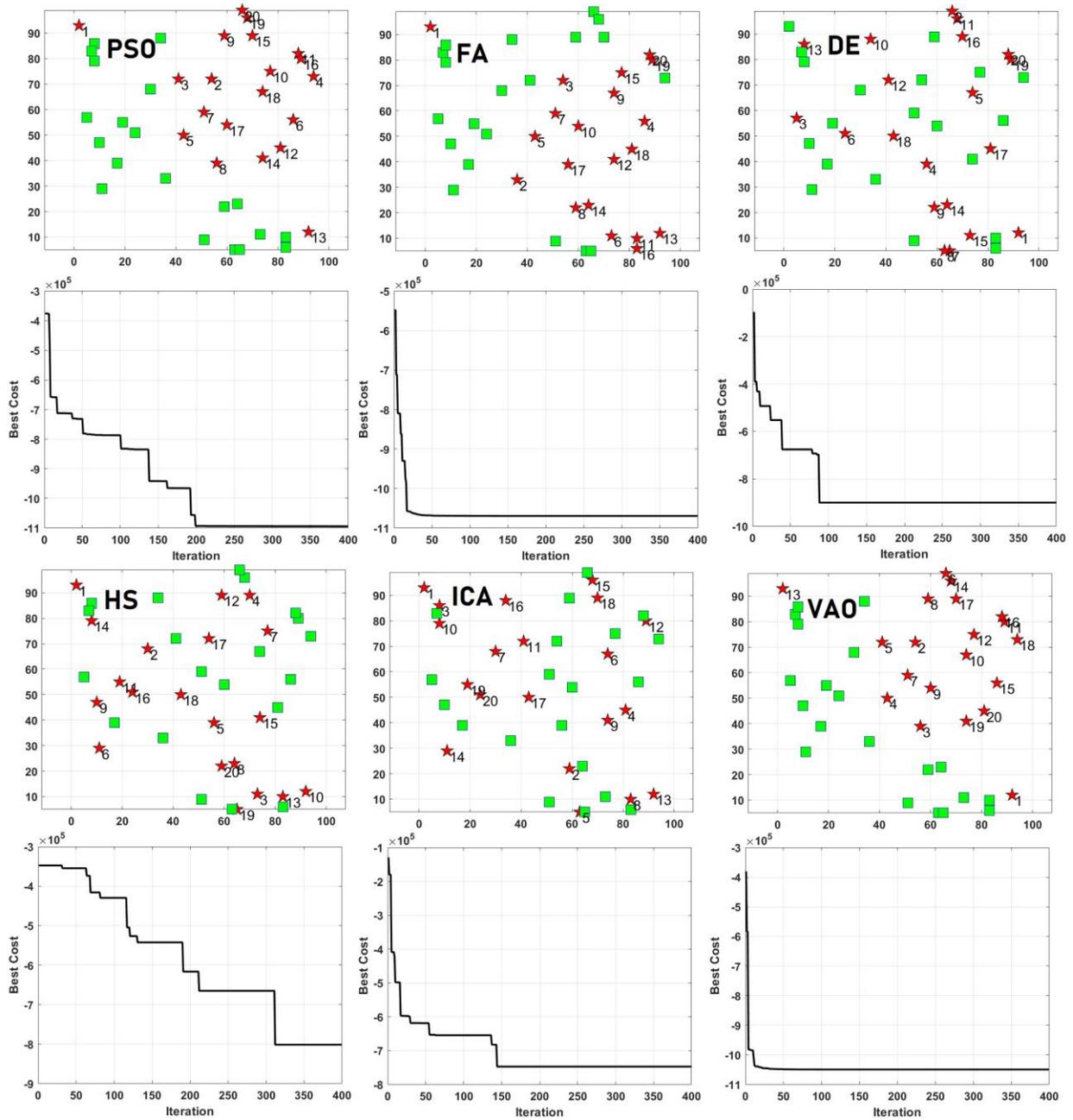

*Figure 13.* FA, PSO, DE, HS, ICA, and VAO algorithm solutions on the same data for the QAP optimization problem (400 iterations and 50 populations).

Table 7

*Comparison results of all algorithms on QAP problem with same parameters*

|  |  | PSO | FA | DE | HS | ICA | VAO |
|---|---|---|---|---|---|---|---|
| **IS PROBLEM SOLVED?** |  | ✓ | ✓ | ✗ | ✗ | ✗ | ✓ |
| **COST VALUE** | Avg | -9.88e+05 | -1.06e+06 | -9.00e+05 | -8.01e+05 | -7.47e+05 | -1.04e+06 |
|  | Std | -8.54e+03 | -1.26e+03 | -8.97e+03 | -7.02e+03 | -6.79e+03 | -9.70e+03 |
| **RUN TIME** | Avg | 3.958 (s) | 10.517 (s) | 1.484 (s) | 1.179 (s) | 1.736 (s) | 12.480 (s) |
|  | Std | 0.44 | 1.25 | 0.39 | 0.41 | 0.35 | 1.11 |

Also, the proposed algorithm is applied to the clustering problem which is an unsupervised machine learning task and compared to other nature-inspired and conventional methods. Clustering (Mousavi *et al.*, 2017b) is the process of grouping similar objects into clusters based on distance factors in multiple dimensions or features and has widespread use in the field of pattern recognition. Here, the Iris benchmark dataset (Fisher, 1936) is used as the experiment data. With three clusters, 30 individuals as population size, and over 300 iterations, the proposed VAO for the clustering method is compared to PSO for clustering (Zhang & Peng, 2022), FA for clustering (Žunić *et al.*, 2021), DE for clustering (Kwedlo, 2011), HS for clustering (Amiri et al., 2010), and ICA for clustering (Niknam et al., 2011). In addition, for comparative purposes, the well-known K-means (Mousavi, 2019) and Gaussian Mixture Model (GMM) (Mousavi, 2019) clustering techniques are applied to the same data. Figure 15 depicts this comparison of features one and four of the Iris dataset. Figure 8 shows the comparison results of the clustering problem on the iris dataset by all algorithms. All algorithms could successfully cluster the iris dataset. However best performance regarding cost value goes for VAO and the worst performance regarding to runtime goes for the VAO algorithm, again. It has to be mentioned that K-means and GMM are faster than optimization algorithms as there is no iteration in their algorithms.

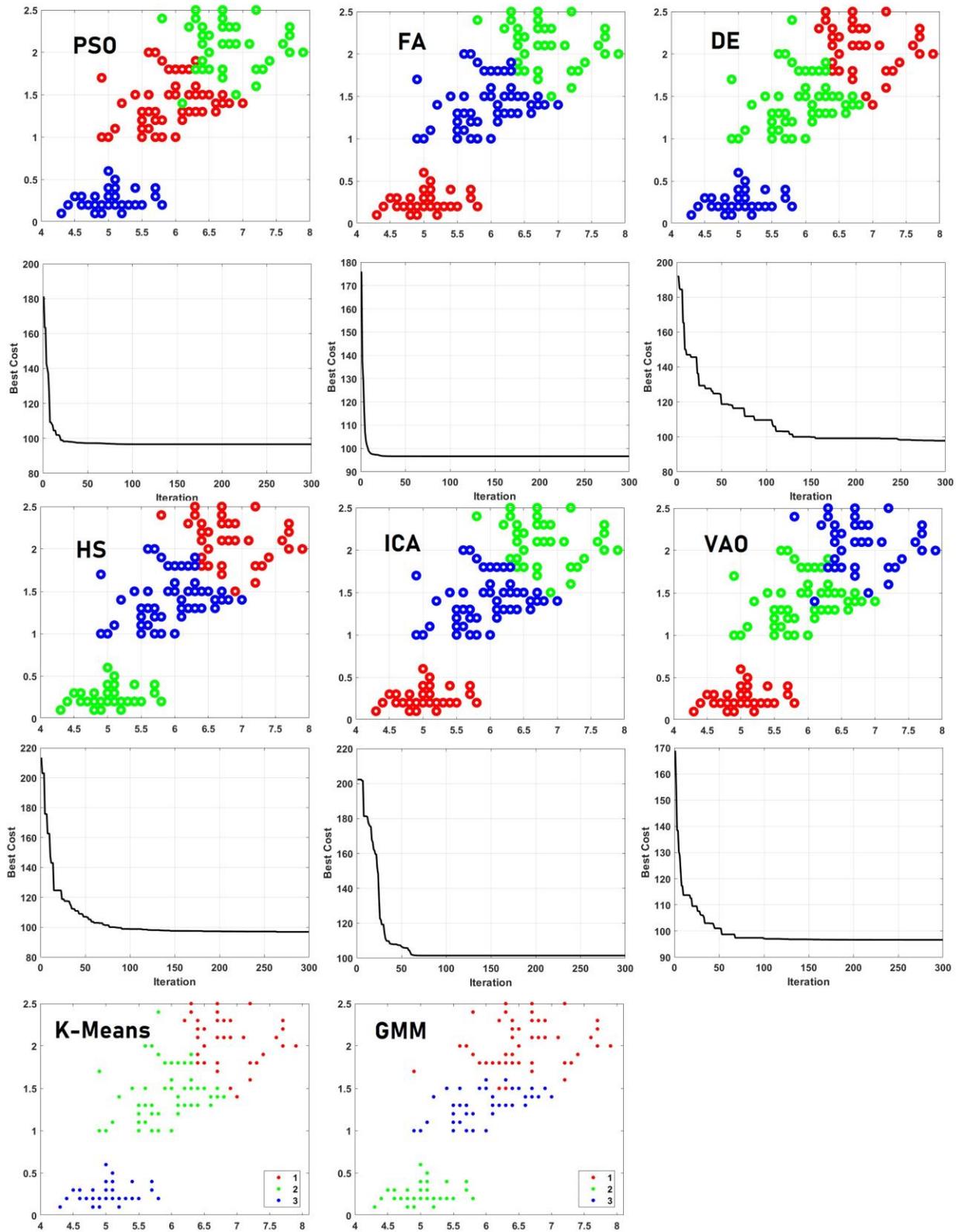

*Figure 15.* FA, PSO, DE, HS, ICA, K-Means, GMM, and VAO algorithm solutions on the Iris data (features 1 and 4) for the Clustering optimization problem (300 iterations and 30 populations).

Table 8

*Comparison results of all algorithms on Clustering problem with same parameters*

| | | PSO | FA | DE | HS | ICA | K-MEANS | GMM | VAO |
|---|---|---|---|---|---|---|---|---|---|
| **IS PROBLEM SOLVED?** | | ✓ | ✓ | ✓ | ✓ | ✓ | ✓ | | ✓ |
| **COST** | Avg | 96.67 | 96.66 | 97.01 | 96.95 | 101.43 | None | None | 96.65 |
| **VALUE** | Std | 2.99 | 4.90 | 3.26 | 3.67 | 4.59 | - | - | 3.01 |
| **RUN TIME** | Avg | 1.895 (s) | 17.368 (s) | 1.645 (s) | 1.415 (s) | 1.740 (s) | 0.10 | 0.15 | 17.954 (s) |
| | Std | 0.12 | 2.39 | 0.29 | 0.24 | 0.37 | 0.01 | 0.01 | 2.50 |

Another task that could be defined as an optimization problem is Feature Selection (FS) task (Mistry, Kamlesh, *et al*, 2016). In feature selection, we are dealing with the Number of Features of "NF", the weight of the feature or "w" and Mean Square Error (MSE) (Karunasingha, Dulakshi, 2022) which should be minimized to select the feature. By reducing the number of features, computational time, decreases, and outliers eliminate from the data. Also, if $x_i$ is the value of NF then, $\hat{x_i}$ would be selected features out of NF. So, considering the number of features entering the system, "y' would be the output, and "t" would be the target. In order to calculate the final error, $e_i$ needs to be calculated which is $t_i - y_i$. So, the final error is $\min MSE = \frac{1}{n} \sum_{i=1}^{n} e_i^2 + w*NF$. This goes for all features and finally, those features with the lowest MSE will be selected. In combination with VAO and feature selection, each feature vector is considered a plant with a different cost function. Those plants which could fit into the final iteration would be selected alongside their related features with lowers error as mentioned. Here one dataset of Statlog (Quinlan, J. Ross, 1987) dataset is employed to validate the task. Statlog is a financial dataset consisting of 690 instances in 14 features belonging to regressions and classification (Dezfoulian et al., 2016) areas. Here classification accuracy for all data, half of the features and a quarter of features is reported using K-Nearest Neighbor (K-NN) (Nugrahaeni, Ratna Astuti, 2016) and Support Vector Machine (SVM) (Nugrahaeni, Ratna Astuti, 2016) classification algorithms and compared with other nature-inspired features selection techniques. Nature-inspired FS should be able to select the best features with a minimum amount of data loss with the same parameters compared with other similar methods. VAO FS is compared with PSO FS (Mistry, Kamlesh, *et al*, 2016), FA FS (Emary, Eid, *et al*, 2015), DE for FS (Khushaba et al., 2008), HS for FS (Diao et al., 2012), ICA for FS (Rad et al., 2012), (Principal Component Analysis (PCA) FS (Saidi, Fatiha, *et al*, 2022) and Lasso FS (Zhang, Huaqing, *et al*, 2019). Table 9 represents SVM and KNN classification results on mentioned datasets with all features, half of the features, and a quarter of the features for all feature selection algorithms. Also, Figure 16 depicts PSO FS, FA FS, DE FS, HS FS, ICA FS, and VAO FS cost functions during 50 iterations. All algorithms succeeded to solve the problem by selecting impactful features. However, the HS algorithm had the best performance and PCA the worst. Table 10 presents, a comparison of cost and run time results for FS problem by all algorithms on half of features.

Table 9

*Classification results for 100 %, 50 % and 25% of features using different feature selection algorithms and methods using SVM and KNN classifiers on Statlog datasets.*

| DATASET | PARAMETERS | PSO FS | FA FS | DE FS | HS FS | ICA FS | PCA FS | LASSO FS | VAO FS |
|---|---|---|---|---|---|---|---|---|---|
| **STATLOG (AUSTRALIAN CREDIT APPROVAL)** | SVM All: 14 Features | 86.4 % | 86.4 % | 86.4 % | 86.4 % | 86.4 % | 86.4 % | 86.4 % | 86.4 % |
| | SVM Half: 7 Features | 91.1 % | 93.8 % | 88.5 % | 94.9 % | 91.8 % | 80.9 % | 84.7 % | 91.2 % |
| | SVM Quarter: 4 Features | 86.2 % | 85.2 % | 83.7 % | 87.2 % | 85.3 % | 74.7 % | 82.5 % | 85.7 % |
| | KNN All: 14 Features | 86.7 % | 86.7 % | 86.7 % | 86.7 % | 86.7 % | 86.7 % | 86.7 % | 86.7 % |
| | KNN Half: 7 Features | 88.2 % | 88.8 % | 85.1 % | 89.1 % | 87.9 % | 81.5 % | 87.9 % | 89.7 % |
| | KNN Quarter: 4 Features | 86.3 % | 87.6 % | 84.2 % | 87.3 % | 85.4 % | 78.1 % | 85.8 % | 86.4 % |

Table 10

*Comparison cost and run time results of all algorithms on FS problem with same parameters (Half Features)*

| | | PSO | FA | DE | HS | ICA | PCA | LASSO | VAO |
|---|---|---|---|---|---|---|---|---|---|
| **IS PROBLEM SOLVED?** | | ✓ | ✓ | ✓ | ✓ | ✓ | | | ✓ |
| **COST** | Avg | 0.083 | 0.090 | 0.090 | 0.079 | 0.085 | None | None | 0.083 |
| **VALUE** | Std | 0.005 | 0.004 | 0.006 | 0.003 | 0.004 | - | - | 0.003 |
| **RUN TIME** | Avg | 60.485 | 61.631 | 43.659 | 59.448 | 59.859 | 2.250 | 10.341 | 59.328 |
| | Std | 1.52 | 1.69 | 1.77 | 1.20 | 1.58 | 0.50 | 0.95 | 1.01 |

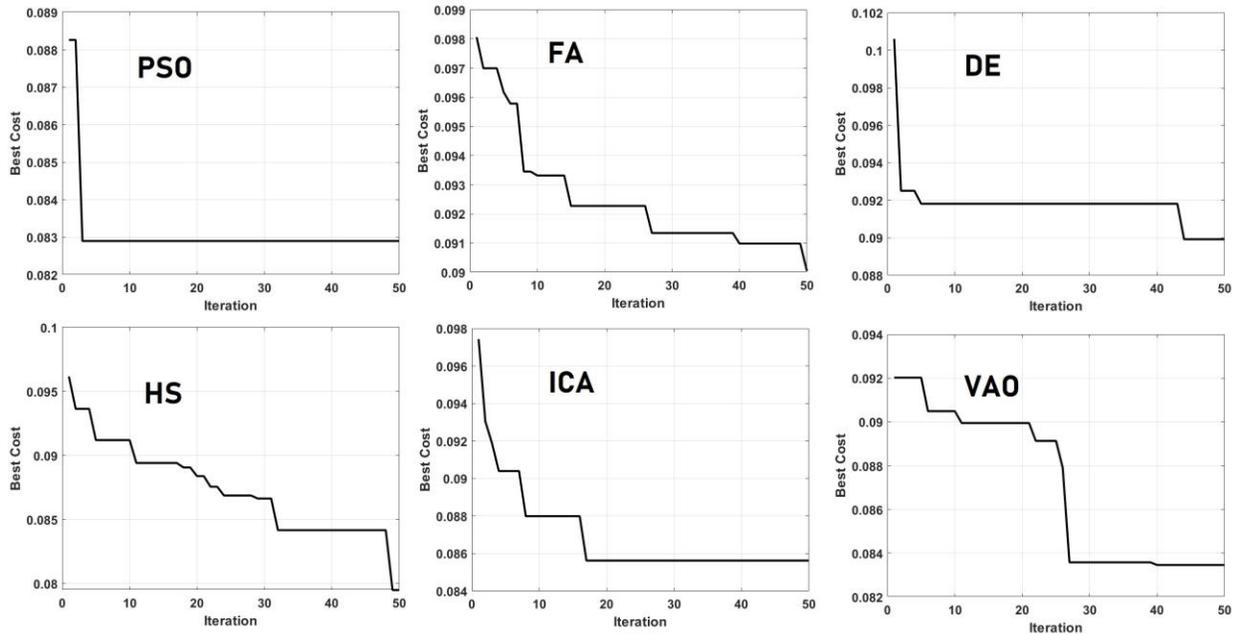

*Figure 16.* PSO FS, FA FS, DE FS, HS FS, ICA FS, and VAO FS cost functions during 50 iterations of runtime with same number of populations (10).

In order to determine the statistical relation power between dependent and independent variables, a regression task gets employed and could be defined as an optimization problem to solve by reducing error for inputs and targets in the data. This will be achieved by improving the weights and biases of a trained algorithm during iterations. Here, the shallow neural network of One Step Secant (OSS) (Upadhyay, 2013) is used for creating a basic model with 9 neurons. The Daily Demand Forecasting Order dataset (Ferreira, R. Pinto, *et al*, 2016) is used for this experiment. This dataset consists of 60 instances in 13 features. Proposed VAO regression is compared with PSO regression (Semero et al., 2018), FA regression (Moazenzadeh et al., 2018), DE regression (Wang et al., 2012), HS regression (Choudhary et al., 2018), ICA regression, and (Azadeh et al., 2016) on the dataset and validated using MSE, Root Mean Square Error (RMSE) (Karunasingha, Dulakshi, 2022) and Correlation Coefficient (CC) (Asuero, Agustin Garcia, *et al*, 2006) performance metrics and the result are presented in Table 11. All algorithms run on 50 populations and 150 iterations. Also, the comparison cost and run time results of all algorithms are represented in Table 12. Figure 17 depicts regression results alongside related cost values. All algorithms succeeded to solve the regression problem in which PSO and VAO, and FA returned better performance in all factors. On the other hand, DE was the weakest algorithm in all factors except the CC factor which HS achieved the lowest value of CC. It has to be mentioned that ICA returned medium-quality performance.

Table 11

*MSE, RMSE and CC performance metrics results for different regression algorithms as a comparison on daily demand dataset.*

|  |  | PSO | FA | DE | HS | ICA | VAO |
|---|---|---|---|---|---|---|---|
| **MSE** | Avg | 0.0016 | 0.0027 | 0.1633 | 0.0391 | 0.0432 | 0.0027 |
|  | Std | 0.0002 | 0.0005 | 0.0224 | 0.0035 | 0.0049 | 0.0002 |
| **RMSE** | Avg | 0.040 | 0.052 | 0.404 | 0.197 | 0.207 | 0.051 |
|  | Std | 0.005 | 0.007 | 0.069 | 0.061 | 0.037 | 0.004 |
| **CC** | Avg | 0.974 | 0.958 | 0.635 | 0.547 | 0.725 | 0.960 |
|  | Std | 0.025 | 0.064 | 0.029 | 0.047 | 0.063 | 0.029 |

Table 12

*Comparison results of all algorithms on Regression problem with same parameters*

|  |  | PSO | FA | DE | HS | ICA | VAO |
|---|---|---|---|---|---|---|---|
| **IS PROBLEM SOLVED?** |  | ✓ | ✓ | ✓ | ✓ | ✓ | ✓ |
| **COST VALUE** | Avg | 0.089 | 0.082 | 4.906 | 1.174 | 1.297 | 0.081 |
|  | Std | 0.004 | 0.006 | 0.754 | 0.096 | 0.070 | 0.005 |
| **RUN TIME** | Avg | 80.120 (s) | 270.362 (s) | 79.923 (s) | 43.898 (s) | 84.561 (s) | 810.263 (s) |
|  | Std | 3.65 | 9.211 | 4.60 | 2.55 | 7.64 | 12.14 |

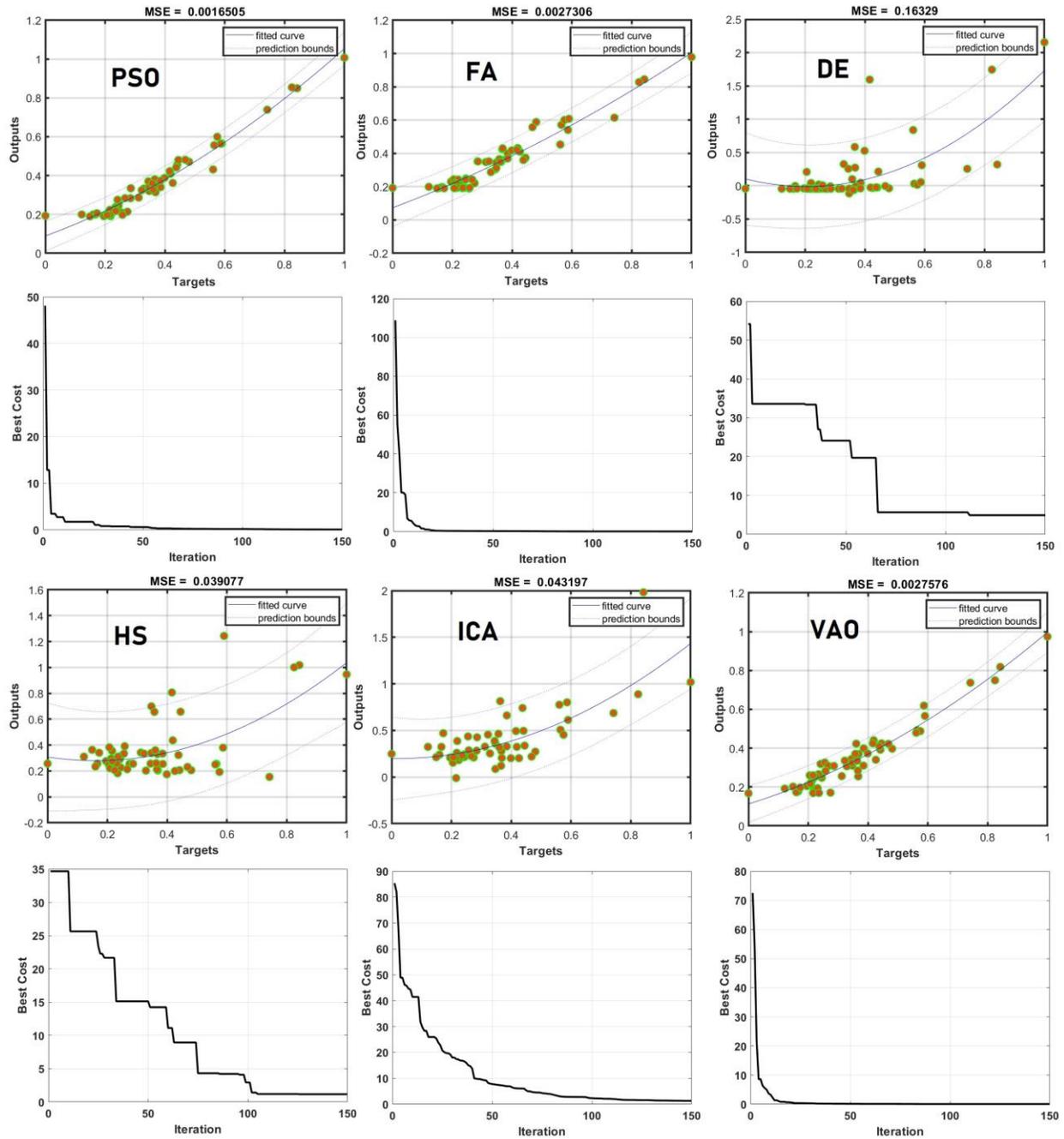

*Figure 17*. Comparison results for regression problem by PSO, FA, DE, HS, ICA, and VAO algorithms (150 iterations and 50 populations).

One of the important subjects in electrical engineering and operational research is Economic Dispatching (ED) in which Power Demand (PD) value must be satisfied by multiple power plants or power generators in a way that the lowest possible cost meets. Each power generators produce a different amount of power according to its technology and size which makes Power Total (PT). Also, there is a bit of Power Loss (PL) in satisfying the total power which should be considered. In this experiment, there are six power generators of P with various minimum and maximum power generations. Here min and max powers for all six power plants are as follows Min= [100 50 80 50 50 50] and, Max= [500 200 300 150 200 120]. The Power Demand (PD) is 1100 megawatts which must be satisfied by the sum of the generated powers using power generators with a very low error. Table 13 shows the performance of all algorithms for the ED problem (Sheta et al., 2020) and for each power generator to satisfy 1100-megawatt power alongside with final power loss and error. Error is total generated power minus power loss during power generation minus initial demanded power by the problem. Also, the final cost value and runtime for each algorithm are reported in Table 14. Figure 18 depicts cost value comparison for all algorithms during 200 iterations and 40 populations size. However, all errors are so low and all algorithms solved the problem, PSO had the best performance and HS the worst. It has to be mentioned that, HS was the fastest algorithm and FA the slowest. The proposed VAO algorithm stands in second place by all factors after PSO. Figure 19 represents Table 13 as a 3-D bar plot.

Table 13

*Comparison results of all algorithms on Regression problem with same parameters*

| PD = 1100 | PSO | FA | DE | HS | ICA | VAO |
|---|---|---|---|---|---|---|
| **P1** | 412 | 413 | 408 | 407 | 396 | 413 |
| **P2** | 140 | 140 | 128 | 139 | 140 | 140 |
| **P3** | 245 | 240 | 240 | 240 | 240 | 240 |
| **P4** | 110 | 129 | 147 | 127 | 120 | 128 |
| **P5** | 150 | 134 | 133 | 137 | 122 | 135 |
| **P6** | 67 | 66 | 65 | 72 | 105 | 65 |
| **PT** | 1125 | 1124 | 1123 | 1124 | 1124 | 1124 |
| **POWER LOSS (PL)** | 24.97 | 23.92 | 23.27 | 23.95 | 24.19 | 23.95 |
| **ERROR=PT-PL-PD** | 6.72e-06 | 8.85e-04 | 0.0380 | 0.0904 | 0.0221 | 6.39e-04 |

Table 14

*Comparison results of all algorithms on ED problem with same parameters*

| | | PSO | FA | DE | HS | ICA | VAO |
|---|---|---|---|---|---|---|---|
| **IS PROBLEM SOLVED?** | | ✓ | ✓ | ✓ | ✓ | ✓ | ✓ |
| **COST VALUE** | Avg | 13475 | 13468 | 13476 | 13470 | 13492 | 13467 |
| | Std | 250 | 396 | 412 | 349 | 880 | 197 |
| **RUN TIME** | Avg (s) | 0.979 (s) | 11.448 (s) | 0.991 (s) | 0.785 (s) | 1.432 (s) | 9.094 (s) |
| | Std | 0.125 | 1.536 | 0.188 | 0.149 | 0.223 | 1.103 |

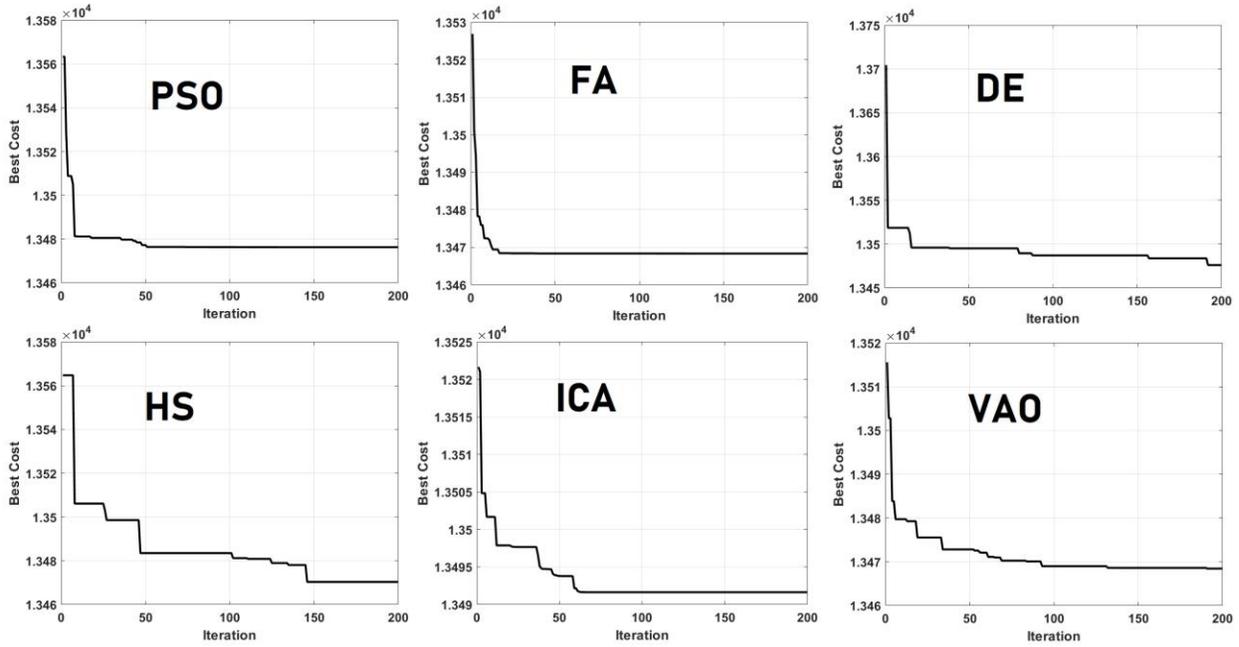

*Figure 18.* Comparison cost value results for ED problem by PSO, FA, DE, HS, ICA, and VAO algorithms (200 iterations and 40 populations).

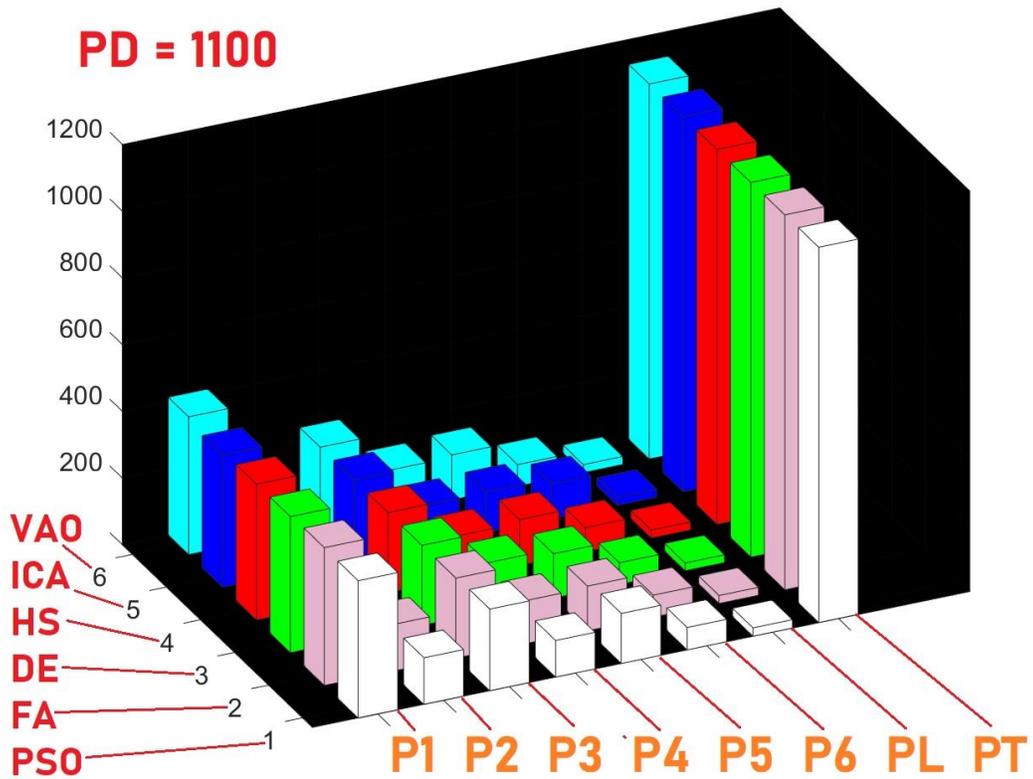

*Figure 19.* 3-D bar plot of Table 13.

A famous problem in time management and operational research is Parallel Machine Scheduling (PMS) which could be solved perfectly by optimization algorithms. In this type of problem, there are multiple machines and tasks with different lengths or complexity. Solve should be set in a way that all tasks be handled by all machines as optimized as possible. Also, there are sets of rules as follow: 1. Each task should be done just by one machine, 2. Each task should be done just once, 3. Starting time of each task is equal to the finishing time of the previous task plus set-up time, 4. Finishing time is equal to starting time plus process time. In this experiment, there are three machines and 10 tasks. The processing time of all tasks is random between 10 and 50 values. Also, set-up time is considered a random number between 3 and 9. We are looking for minimizing the maximum machine completion time or C-Max. Table 15 shows starting data for each machine and task. Table 16 presents the acquired result on the PMS problem by all algorithms with the same parameters on the same data (Table 15). The proposed VAO algorithm performance result comparison with PSO, FA, DE, HS, and ICA (Mokotoff, 2001) in 100 iterations and 20 populations are depicted in Figure 20. In Figure 20, Each green square is a task and inside each square, the number of tasks is presented. Also, the distance between each square represents the set-up time for the next task. All algorithms passed the test, however, the proposed VAO algorithm returned the best performance with C-MAX = 86 and ICA returned the worst performance with C-Max = 112.

Table 15

*PMS initial data for Machines (M) and Tasks (T)*

|      | T1 | T2 | T3 | T4 | T5 | T6 | T7 | T8 | T9 | T10 |
|------|----|----|----|----|----|----|----|----|----|-----|
| **M1** | 23 | 49 | 22 | 11 | 41 | 28 | 47 | 37 | 50 | 11 |
| **M2** | 24 | 39 | 50 | 22 | 34 | 31 | 25 | 11 | 41 | 15 |
| **M3** | 47 | 42 | 29 | 50 | 27 | 47 | 14 | 15 | 28 | 10 |

Table 16

*Comparison results of all algorithms on PMS problem with same parameters*

|      |      | PSO | FA | DE | HS | ICA | VAO |
|------|------|-----|----|----|----|-----|-----|
| **IS PROBLEM SOLVED?** |  | ✓ | ✓ | ✓ | ✓ | ✓ | ✓ |
| **COST VALUE** | Avg | 94 | 93 | 90 | 95 | 112 | 86 |
| **OR C-MAX** | Std | 6 | 2 | 4 | 4 | 5 | 3 |
| **RUN TIME** | Avg (s) | 0.630 (s) | 0.612 (s) | 0.644 (s) | 0.537 (s) | 0.715 (s) | 0.678 (s) |
|  | Std | 0.124 | 0.096 | 0.159 | 0.087 | 0.143 | 0.118 |

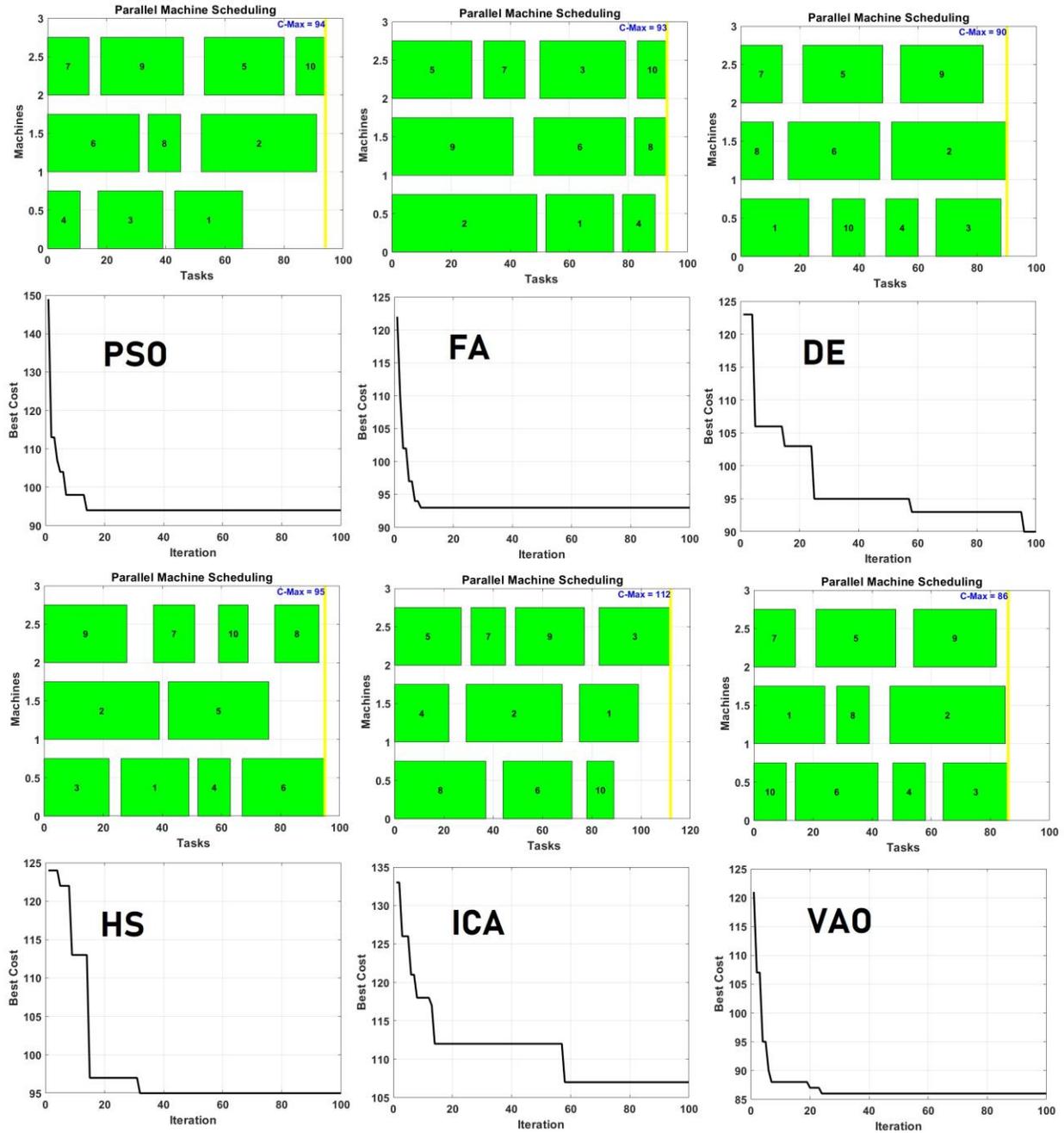

*Figure 20.* Comparison results for PMS problem by PSO, FA, DE, HS, ICA, and VAO algorithms over 100 iterations with 20 population.

Image quantization is compressing a range of color values into a single value which leads to reducing image size and increasing compatibility with high ranges of devices. This could be done by clustering colors and selecting center clusters as a color threshold. Depending on initial color thresholds or clusters, the final image's quality and number of used colors determine. The test image is a benchmark baboon image in the size of 256*256 which goes under experiment with 5, and 12 color thresholds or clusters. Figure 21 shows the baboon test image alongside its related image histogram with 64 and 32 bins. Table 17 shows comparison results for all algorithms on IQ problems as cost value and runtime. Figure 22 depicts the acquired results of all algorithms on IQ problems. All algorithms passed the test in which DE and ICA returned poor performance but high speed. On the other hand, FA and VAO returned the strongest performance but with low speed.

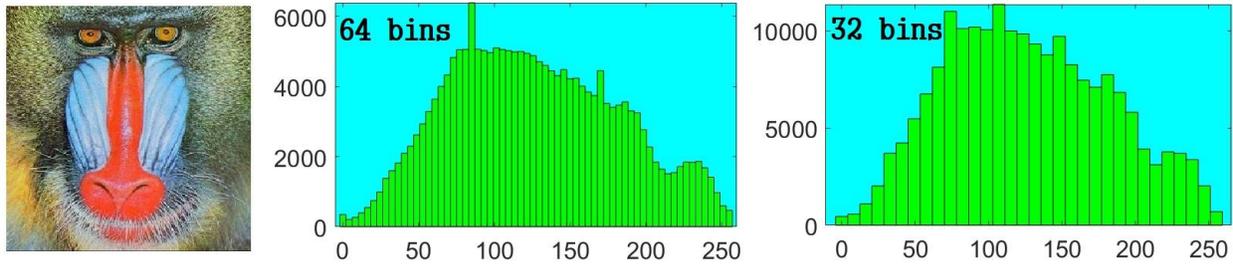

*Figure 21.* Baboon test image and its related image histogram in 64 and 32 bins.

Table 17

*Comparison results of all algorithms on IQ problem with same parameters on 12 thresholds*

|  |  | PSO | FA | DE | HS | ICA | VAO |
|---|---|---|---|---|---|---|---|
| **IS PROBLEM SOLVED?** |  | ✓ | ✓ | ✓ | ✓ | ✓ | ✓ |
| **COST VALUE** | Avg | 8837 | 6667 | 9873 | 8120 | 9667 | 6426 |
|  | Std | 254 | 366 | 473 | 349 | 712 | 127 |
| **RUN TIME** | Avg (s) | 3.477 (s) | 10.542 (s) | 2.964 (s) | 2.889 (s) | 2.566 (s) | 10.633 (s) |
|  | Std | 0.527 | 1.930 | 0.366 | 0.810 | 0.428 | 1.509 |

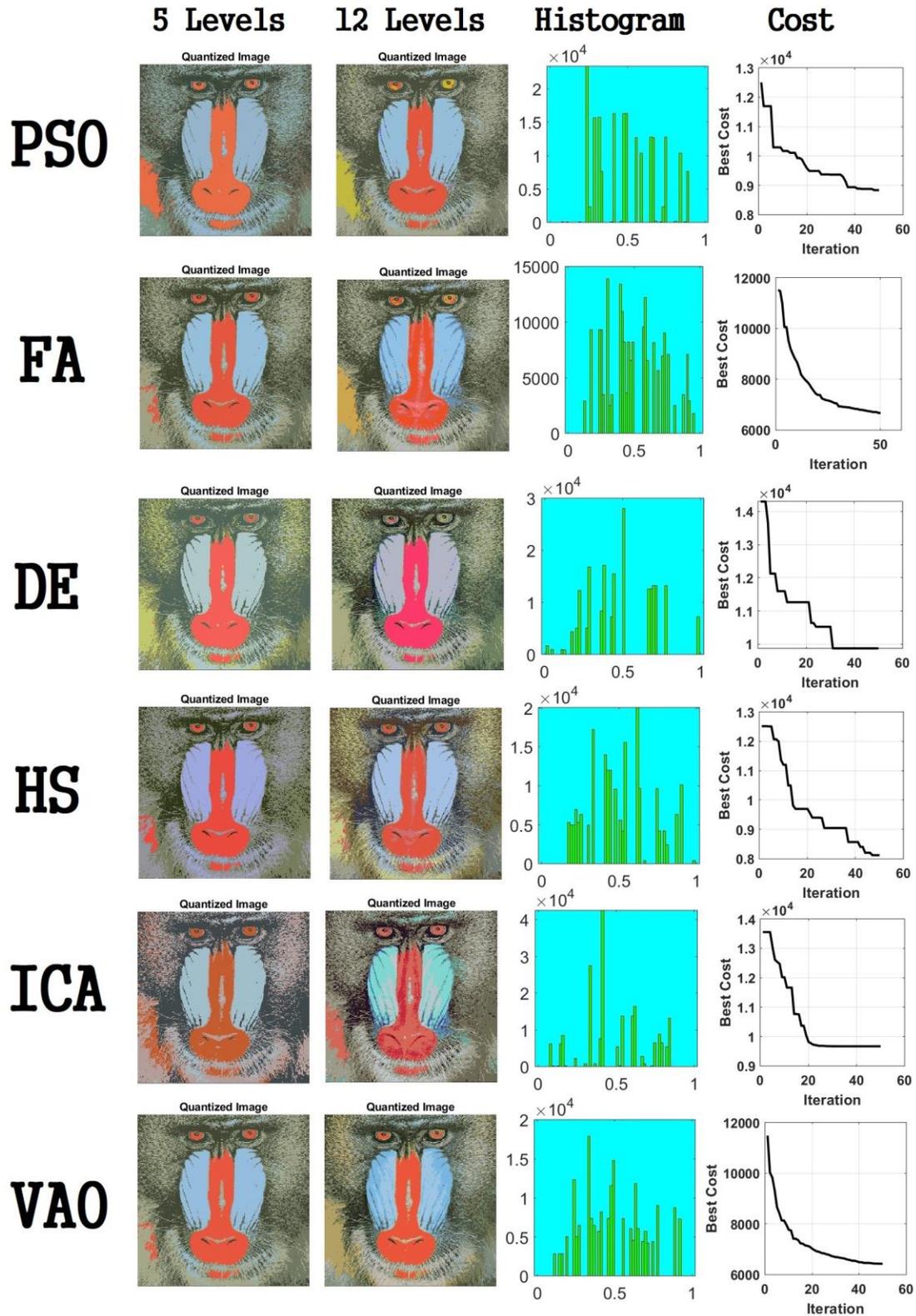

*Figure 22.* Acquired results of different algorithms on IQ problem. Image histogram belongs to 12 levels (50 iterations and 10 populations).

Image Segmentation (IS) is a subset of image processing (Mousavi and Kharazi., 2017) and means dividing a digital image into multiple segments or objects to achieve something meaningful and easier to analyze. In image segmentation, pixels with similar characteristics are categorized into various groups or segments each segment indicates the specific part of the image based on texture, intensity, and color for a better understanding of human eyes and experts. Similar to the IQ problem, the same technique of clustering applies to segment the image. It starts with converting the image matrix into an image vector and improving the K-means clustering core during iterations to achieve cluster centers. Then, each cluster center considers a segment and is presented with a unique color. In this experiment. Monarch benchmark test image is employed. We extracted the butterfly and removed the background to increase the visibility and comparison power. Figure 23 illustrates the original monarch image alongside with extracted butterfly image for the experiment and the corresponding image histogram of extracted butterfly with 32 bins. Table 18 shows returned comparison results for IS problem by all algorithms with 2 segments. Performance metrics are Accuracy, Precision, Recall, F-Score, Matthews Correlation Coefficient (MCC), Peak Signal Noise Ratio (PSNR), and Structural Similarity Index Measure (SSIM) (Wang et al., 2020) which shows the relation between Ground Truth (GT) and returned binary images. Also, Table 19 presents the cost and runtime comparison of all algorithms on IS problem with the same parameters on 2 segments over 40 iterations and 2 populations. Figure 24 depicts the acquired results of IS problem on monarch images by different algorithms over 40 iterations and 2 populations. PSO, FA, and VAO returned high-quality results. However, DE, HS, and ICA showed a medium to proper level performance.

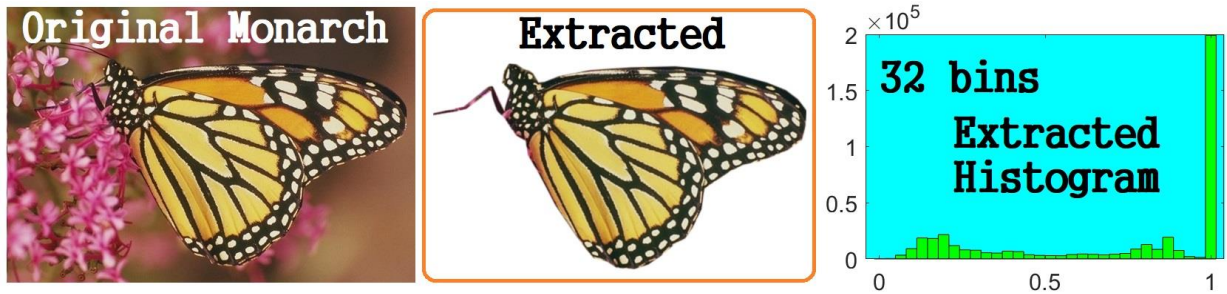

*Figure 23.* From left to right: Monarch test image, extracted butterfly for segmentation, and image histogram of extracted butterfly.

Table 18

*Comparison performance of all algorithms on monarch test image with 2 segments (segmented VS ground truth)*

|  |  | PSO | FA | DE | HS | ICA | VAO |
|---|---|---|---|---|---|---|---|
| **ACCURACY** | Avg | 0.999 | 0.998 | 0.992 | 0.992 | 0.995 | 0.999 |
|  | Std | 0.002 | 0.002 | 0.004 | 0.006 | 0.005 | 0.003 |
| **PRECISION** | Avg | 1 | 1 | 0.998 | 0.974 | 1 | 1 |
|  | Std | 0.001 | 0.002 | 0.001 | 0.009 | 0.002 | 0.004 |
| **RECALL** | Avg | 0.999 | 0.994 | 0.972 | 1 | 0.984 | 1 |
|  | Std | 0.003 | 0.002 | 0.008 | 0.003 | 0.009 | 0.002 |
| **F-SCORE** | Avg | 0.998 | 0.997 | 0.985 | 0.987 | 0.992 | 0.997 |
|  | Std | 0.003 | 0.005 | 0.007 | 0.008 | 0.004 | 0.003 |
| **MCC** | Avg | 0.997 | 0.996 | 0.980 | 0.982 | 0.989 | 0.998 |
|  | Std | 0.005 | 0.001 | 0.006 | 0.007 | 0.006 | 0.004 |
| **PSNR** | Avg | 48.46 | 28.18 | 21.02 | 21.32 | 23.67 | 46.11 |
|  | Std | 4.12 | 6.19 | 5.98 | 3.27 | 7.02 | 5.10 |
| **SSIM** | Avg | 0.998 | 0.989 | 0.952 | 0.958 | 0.972 | 0.998 |
|  | Std | 0.004 | 0.006 | 0.009 | 0.008 | 0.010 | 0.002 |

Table 19

*Cost and runtime comparison results of all algorithms on IS problem with same parameters on 2 segments*

| | | PSO | FA | DE | HS | ICA | VAO |
|---|---|---|---|---|---|---|---|
| **IS PROBLEM SOLVED?** | | ✓ | ✓ | ✓ | ✓ | ✓ | ✓ |
| **COST VALUE** | Avg | 13584 | 13852 | 15484 | 14551 | 14380 | 13581 |
| | Std | 22 | 19 | 35 | 31 | 42 | 11 |
| **RUN TIME** | Avg (s) | 1.005 | 1.043 | 1.171 | 2.094 | 1.147 | 1.054 |
| | Std | 0.009 | 0.004 | 0.010 | 0.053 | 0.033 | 0.006 |

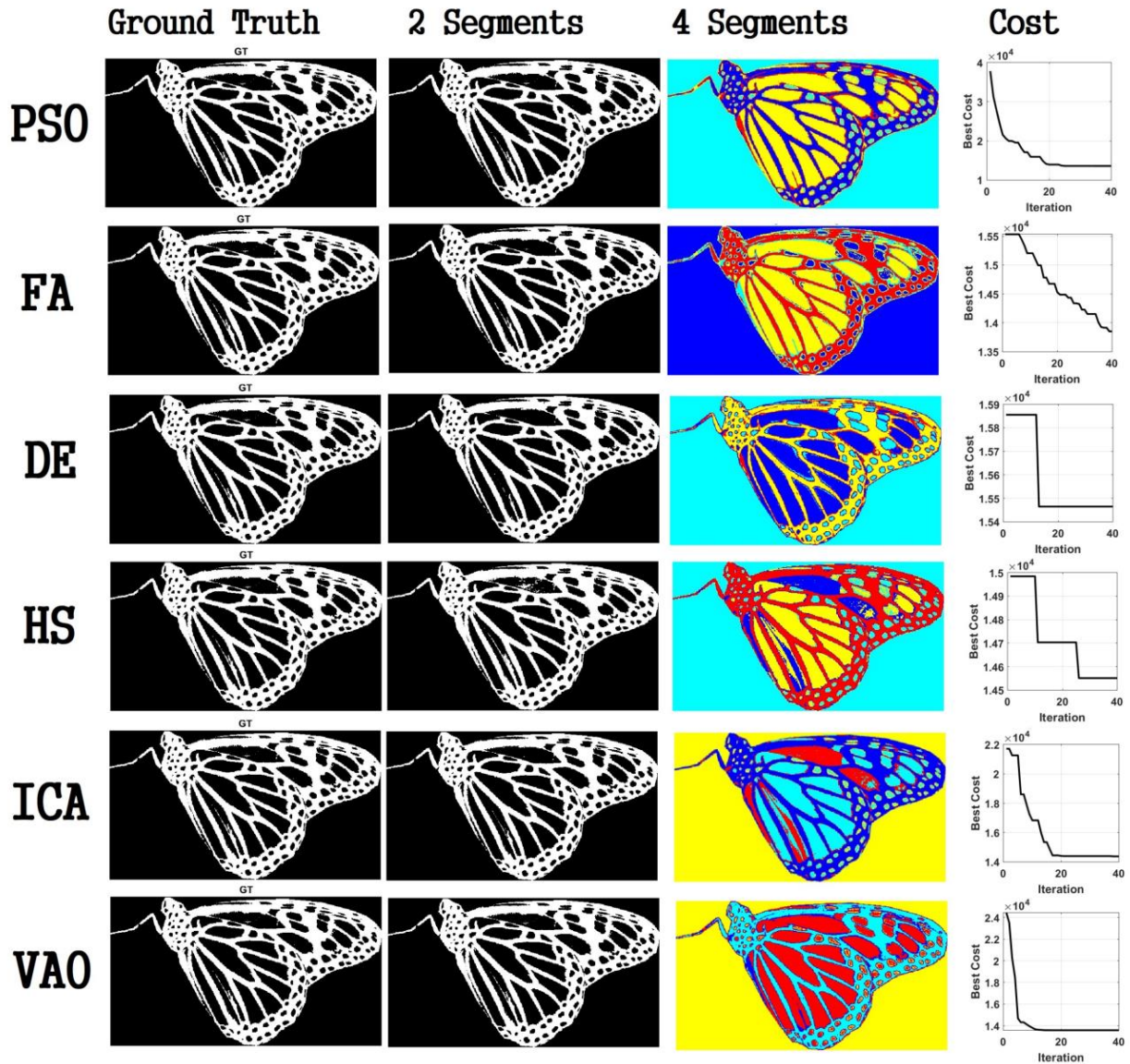

*Figure 24.* Segmentation by all algorithms on monarch test image. From left to right: algorithm name, ground truth image, segmentation by 2 segments, segmentation by 4 segments, and cost value over 40 iteration and 2 populations.

## 5. CONCLUSIONS AND FUTURE RESEARCH AVENUES

Nowadays, the vast majority of research experiment fields involve some type of optimization problem of varied complexity, and nature-inspired optimization approaches could effectively manage this complexity. The proposed VAO algorithm is influenced by the plant life cycle, which is rare but robust and optimized. The VAO demonstrated better convergence than the majority of other nature-inspired benchmark algorithms.

In addition, the VAO algorithm could handle nearly all forms of unimodal and multimodal optimization functions with varying landscape shapes for both single- and multi-objective types. The suggested approach demonstrated robustness, precision, a good convergence rate, and great general performance on 10 distinct optimization problems. Not only was it superior to other well-known algorithms, but it also placed second overall across all tests.

Future research will focus on applying the VAO algorithm to other optimization tasks, such as optimal path planning, portfolio optimization, and facility layout design. It is also suggested that VAO and fuzzy logic be combined for supervised learning purposes or even feature selection in pattern recognition.

### Code Availability:

The code and data of this study will be posted on GitHub after acceptance of the paper.